\documentclass[8pt, letterpaper, twocolumn]{article}

\usepackage{microtype}
\usepackage{graphicx}
\usepackage{booktabs} %

\usepackage{color}
\usepackage{amsmath}
\usepackage{amssymb}
\usepackage{amsfonts}
\usepackage{multirow, varwidth, rotating}
\usepackage{epsfig}
\usepackage{verbatim}
\usepackage{dsfont}
\usepackage{wrapfig}
\makeatletter
\newif\if@restonecol
\makeatother
\usepackage{xr}
\usepackage{booktabs}
\usepackage{comment}
\usepackage{paralist}
\usepackage{flushend}
\usepackage{xcolor}
\usepackage{xspace}

\newcommand{\mb}{\mathbf}

\newcommand{\mbb}{\mathbb}

\usepackage{subcaption}

\DeclareRobustCommand\onedot{\futurelet\@let@token\@onedot}
\def\onedot{.}
\def\eg{\emph{e.g}\onedot} 
\def\ie{\emph{i.e}\onedot}

\newcommand{\argmax}{\operatornamewithlimits{argmax}}

\usepackage{hyperref}

\usepackage{amsmath}
\usepackage{amssymb}
\usepackage{mathtools}
\usepackage{amsthm}

\usepackage{natbib}
\renewcommand{\cite}[1]{\citep{#1}}
\setcitestyle{authoryear,round,citesep={;},aysep={,},yysep={;}}

\usepackage[font=footnotesize]{caption}

\usepackage[margin=0.9in]{geometry}

\usepackage{titlesec}
\titleformat{\section}
  {\normalfont\large\bfseries}{\thesection}{1em}{}
\titleformat{\subsection}
  {\normalfont\normalsize\bfseries}{\thesubsection}{1em}{}
\titleformat{\subsubsection}
  {\normalfont\normalsize\bfseries}{\thesubsubsection}{1em}{}

\usepackage[capitalize,noabbrev]{cleveref}

\theoremstyle{plain}

\theoremstyle{definition}

\theoremstyle{remark}

\usepackage[textsize=tiny]{todonotes}

\newcommand{\ourMethodName}{MSG$^2$}
\newcommand{\ourGraph}{\ourMethodName-noSSP}
\newcommand{\ourGraphFull}{\ourMethodName~without Subtask State Prediction}
\newcommand{\ourGraphMetricFull}{Strict Partial Order Consistency}
\newcommand{\ourGraphMetric}{SPOC}
\newcommand{\ourDataMetric}{Compatibility}
\newcommand{\ourDataMetricFull}{Compatibility}
\newcommand{\msgip}{MSGI+\xspace}

\newcommand{\maintask}{perform CPR}
\newcommand{\ourTitle}{Multimodal Subtask Graph Generation from Instructional Videos}
\newcommand{\statea}{\textit{not started}}
\newcommand{\stateb}{\textit{in progress}}
\newcommand{\statec}{\textit{completed}}
\newcommand{\task}[1]{\textit{#1}}
\newcommand{\subtask}[1]{\texttt{#1}}
\newcommand{\andorgraph}[1]{{\footnotesize\textsf{#1}}}
\newcommand{\tokensymbol}[1]{{\footnotesize\textsf{#1}}}

\newcommand{\supplementary}[1]{#1}

\def\pfuncarg#1{f_{#1}}
\def\pfunc{\pfuncarg{n}}

\def\ilpdata{\mathcal{D}}
\def\ind{\mathbb{I}}

\def\videoset{\mb{V}_\tau}
\def\frameset{\mb{F}}
\def\frame{\mb{f}}
\def\transtextset{\mb{X}}
\def\transtext{\mb{x}}
\def\hiddenset{\mb{H}}
\def\stateset{\mb{S}}
\def\statevec{\mb{s}}
\def\statescalar{s}

\begin{document}

\title{\ourTitle}
\author{
    {\normalsize Yunseok Jang$^{*\dag}$, Sungryull Sohn$^{*\ddag}$, Lajanugen Logeswaran$^{\ddag}$, Tiange Luo$^{\dag}$, Moontae Lee$^{\ddag}$, Honglak Lee$^{\dag\ddag}$} \\[0.3em]
    {\normalsize $^\dag$University of Michigan, Ann Arbor\hspace{1em} $^\ddag$LG AI Research}
}
\date{}
\maketitle
\def\thefootnote{*}\footnotetext{Equal contribution}\def\thefootnote{\arabic{footnote}}

\begin{abstract}

Real-world tasks consist of multiple inter-dependent subtasks (\eg, a dirty pan needs to be washed before it can be used for cooking).
In this work, we aim to model the causal dependencies between such subtasks from instructional videos describing the task.
This is a challenging problem since complete information about the world is often inaccessible from videos, which demands robust learning mechanisms to understand the causal structure of events. 
We present Multimodal Subtask Graph Generation (\ourMethodName), an approach that constructs a \emph{Subtask Graph} defining the dependency between a task's subtasks relevant to a task from noisy web videos.
Graphs generated by our multimodal approach are closer to human-annotated graphs compared to prior approaches. 
\ourMethodName~further performs the downstream task of next subtask prediction 85\% and 30\% more accurately than recent video transformer models in the ProceL and CrossTask datasets, respectively.

\end{abstract}

\section{Introduction}
\label{sec:introduction}

Humans perceive the world and coherently comprehend a series of events by integrating multiple sensory inputs. 
When we watch an instructional video about how to \task{\maintask}\footnote{Examples: \url{https://youtu.be/t2LrQGD2y2I}, \url{https://youtu.be/gn8Oj1xcWrs}.} (\ie, a \emph{task}), we subconsciously understand the key events, or \emph{subtasks}, such as \subtask{check breathing} or \subtask{give compression}, involved in the \task{task}.
In addition, we can also reason about the dependencies between these subtasks (\eg, checking for breathing before starting chest compressions), even if such information is not explicitly mentioned in the video.
Although discovering these key subtasks and their dependencies comes naturally to humans through strong commonsense priors, replicating such abilities in machines is challenging.

\begin{figure*}[t]
    \centering
    \vspace{-0.5em}
    \includegraphics[width=0.95\linewidth]{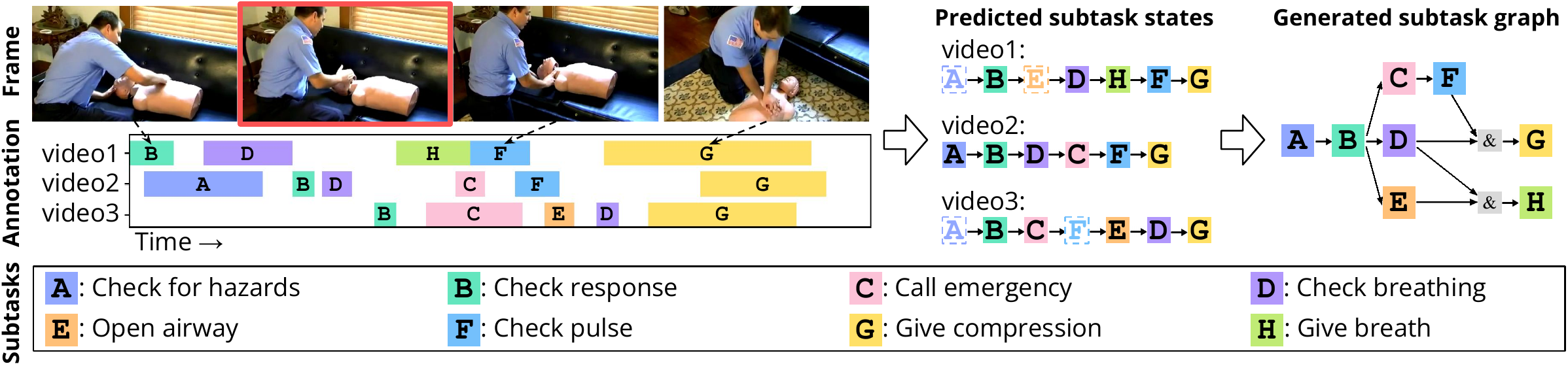}
    \vspace{-0.5em}
    \caption{%
    Given video instances describing a task such as \task{\maintask}, along with noisy subtask annotations, we attempt to extract causal dependencies between the subtasks.
    Data noise presents a major challenge in identifying these dependencies -- (i) subtask \subtask{A} is omitted in videos 1 and 3 (ii) even though subtask \subtask{E} is present in video 1 (frame highlighted in red outline) it is missing from the annotations. 
    We predict such missing steps (dotted boxes) by leveraging visual and text signals (\Cref{subsec:method_subtask_state_prediction}) and generate a subtask graph based on the updated subtask sequences (\Cref{subsec:method_graph_inference}).
    }
    \label{fig:teaser}
    \vspace*{-0.4em}
\end{figure*}

Prior related approaches to learning from videos have largely focused on training video representations useful for downstream tasks from videos and their text transcripts~\cite{miech-iccv19,miech-cvpr20,sun-iccv19,zellers-neurips21} or understanding the structure among key-events in a single video~\cite{elhamifar-iccv19,tang-cvpr19}. 
Different from prior works, we focus on extracting a holistic understanding of a task from \emph{multiple} videos describing the task. 
Learning from multiple videos (as opposed to a single video demonstration) has two key benefits.
First, a task can be accomplished in various ways, and a single video cannot capture such diversity.
Second, instructional videos are noisy in nature (\ie, due to steps being omitted or rearranged. An example is in~\Cref{fig:teaser}) and learning from multiple videos can help improve robustness to such noise.

Our goal is to infer subtask dependencies from multi-modal video-text data and generate a concise graph representation of the subtasks that shows these dependencies. 
In particular, we focus on instructional videos describing how to perform real-world activities such as \task{jump-starting a car} or \task{replacing an iPhone’s battery}. %
Condensing information in the instructional videos in the form of a graph makes the information more easily accessible (\eg, to human users) \citep{vicol-cvpr18, wang-eccv18, zhao-tpami22}.
Specifically, we consider a subtask graph representation that is more expressive at modeling subtask dependencies (\eg, \andorgraph{AND} ($\&$) nodes allow modeling multiple preconditions) compared to representations considered in the literature such as partial order graphs or dataflow graphs \citep{sakaguchi-emnlpf21,madaan2022language}.
Knowledge about subtask dependencies can further benefit downstream applications such as future prediction and robot planning ~\cite{liu-aaai22,sohn-neurips18,sohn-iclr20}.

Instructional videos present a unique set of challenges in terms of understanding task structure due to data noise. 
Certain key events (or subtasks) in the video may be skipped in favor of brevity (\eg, In~\Cref{fig:teaser}, \subtask{A: check for hazards} is skipped in video 1 and 3 as they assume the events are taking place in a safe space free of any hazards). %
In addition, human annotators fail to annotate certain subtasks (\eg, In~\Cref{fig:teaser}, \subtask{E: open airway} appears in video 1 (frame highlighted in red outline) but is missing in the annotations). 
Some prior works~\cite{sohn-iclr20, liu-aaai22, xu-icra18, hayes-icra16, nejati-icml06} extract subtask dependencies (high-level information such as the precondition of the subtasks) from a noise-free simulation environment.
Such methods designed to work with clean data fail to accurately extract subtask dependencies from noisy real-world videos due to the aforementioned challenges.

To tackle these challenges, we present Multimodal Subtask Graph Generation (\ourMethodName) in order to robustly learn task structure from online real-world instructional videos. %
\ourMethodName~first extracts the high-level subtask information from the subtask annotations in the data.
Then, \ourMethodName~present a multimodal subtask state prediction module that makes use of video and text data to accurately infer the missing subtasks to improve the graph inference (\Cref{subsec:method_subtask_state_prediction}). %
We then adopts a complexity-regularized inductive logic programming (ILP) method to handle the extracted incomplete and noisy subtask information (\Cref{subsec:method_graph_inference}).
Lastly, we demonstrate the utility of~\ourMethodName~by employing predicted subtask graphs in a downstream task of predicting the next subtask in a video. %

In summary, we make the following contributions:
\begin{itemize}
\setlength\itemsep{0.0em}
\item We study the problem of predicting the task structure (\ie, the causal dependencies between the subtasks) from the videos describing how to perform the task.%
\item Our multimodal subtask graph generation (\ourMethodName) approach produces subtask graphs that better resemble human-annotated graphs compared to prior approaches. 
\item Subtask graphs generated using our approach improves the performance by a large margin over prior methods on the next subtask prediction task. %
\end{itemize}

\section{Background}
\label{sec:background}
Our work builds on the subtask graph framework~\citep{sohn-neurips18, sohn-iclr20}, which describes the causal dependency structure of a complex task $\tau$ consisting of $N_\tau$ subtasks. %
Each subtask has a \textbf{precondition} that must be satisfied before the subtask can be completed. 
Precondition describes the causal relationship between subtasks and imposes a constraint on the order in which subtasks can be completed (\eg, a pan must be washed \emph{before} being used for cooking). %
Formally, the precondition is defined as a Boolean expression consisting of Boolean constants (\eg, True or False), Boolean variables and logical connectives (\eg, \andorgraph{AND} ($\&$), \andorgraph{OR} (\( \mid \))). 
For instance, consider an example where the precondition of subtask \subtask{C} is $\pfuncarg{\subtask{C}} = \&(\subtask{A}, \subtask{B})$ (\ie, subtasks \subtask{A} and \subtask{B} must be completed before performing \subtask{C}).
The boolean expression $\pfuncarg{\subtask{C}} = \&(\subtask{A}, \subtask{B})$ can be viewed as a \textbf{graph} with vertices consisting of subtasks and logical operators $V = \{\subtask{A}, \subtask{B}, \subtask{C}, \&\}$ and edges $E=\{\subtask{A} \rightarrow \&$, $\subtask{B} \rightarrow \&$, $\& \rightarrow \subtask{C}\}$ that represent preconditions. %
$\pfuncarg{\subtask{C}}$ can also equivalently be viewed as a \textbf{function} that computes whether the precondition of \subtask{C} is satisfied, given the completion status of subtasks \subtask{A} and \subtask{B}.
For instance, if \subtask{A} has been completed (\ie, \subtask{A} = True\footnote{In an abuse of notation, we use \subtask{A} = True to mean subtask \subtask{A} has been completed.}) and \subtask{B} has not been completed (\ie, \subtask{B} = False), we can infer that the precondition of \subtask{C} is not satisfied: $\pfuncarg{\subtask{C}}(\subtask{A}=\text{True},\subtask{B}=\text{False}) = \text{True}~\&~\text{False} = \text{False}$.
We will use these different views of the precondition (\ie, as a boolean expression, graph or function) interchangeably.
The \textbf{subtask graph} visualizes the preconditions $\pfuncarg{1},\ldots,\pfuncarg{N_\tau}$ of the subtasks (see \Cref{fig:teaser,fig:ilp} for examples).
We note that the subtask graph is one of the most flexible frameworks to represent compositional task structure. It has been adopted in various settings~\cite{sohn2022fast, liu-aaai22, sohn-iclr20} and subsumes other task graph formats~\cite{boutilier-ijcai95, andreas-icml17, sakaguchi-emnlpf21}.

\begin{figure*}[t]
    \centering
    \includegraphics[width=0.95\linewidth]{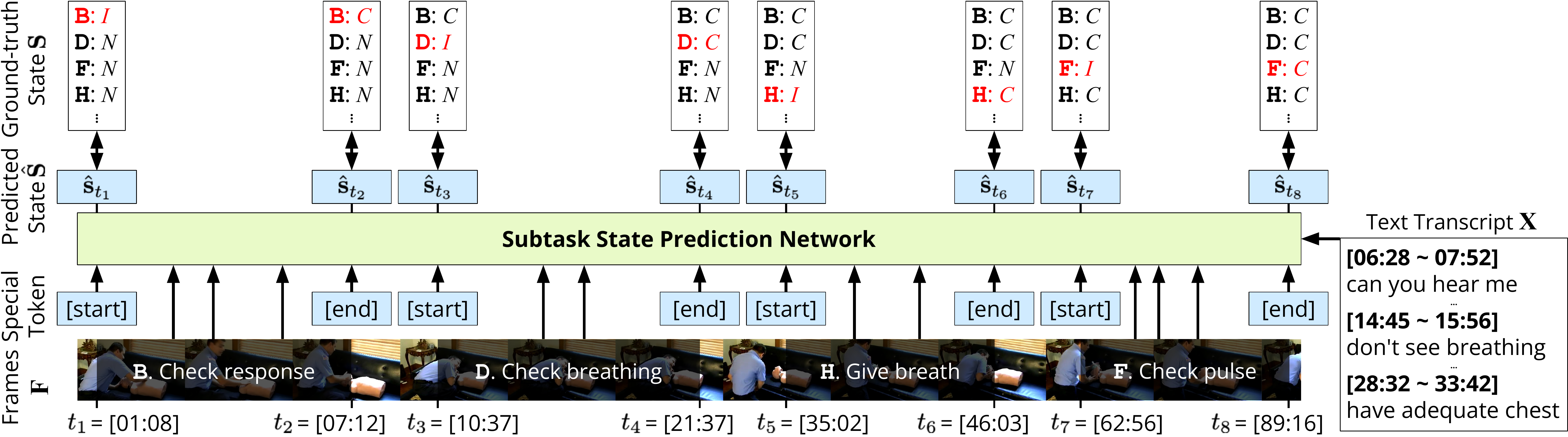}
    \caption{\textbf{Architecture of subtask state prediction module.} 
    We denote subtask state labels as \textit{N} (\statea), \textit{I} (\stateb), \textit{C} (\statec), and labels in red represent ongoing subtask.
    We train the model to predict each subtask's state and use it to predict missing subtask labels in the data, which in turn contributes to more accurate graph generation.
    }
    \label{fig:subtask_state_prediction}
\end{figure*}

\section{Problem Formulation}
\label{sec:problem}

\paragraph{Notation.} %
We use the following convention for notation in the rest of the paper. 
Matrices are denoted by upper-case boldface letters (\eg,~$\hat{\stateset}$), vectors by lower-case boldface letters (\eg,~$\hat{\statevec}$) and scalars by lowercase non-boldface letters (\eg,~$\hat{\statescalar}$).
We use subscripts (\eg,~$\hat{\statevec}_t$) for indexing time %
and array index notation for indexing vectors by subtask (\eg,~$\hat{\statescalar}_t[n]$ denotes state of $n^{\text{th}}$ subtask at time $t$).

\vspace*{-0.1in}
\paragraph{Problem.} We are given a set of instructional videos $\videoset$ (\ie, task instances) describing a task $\tau$ (\eg, \task{\maintask}, \task{jump car}).
Each video consists of video frame $\frameset  = (\frame_1, \frame_2, \ldots, \frame_T)$, text transcript $\transtextset=(\transtext_1, \transtext_2, \ldots, \transtext_T)$, and subtask state labels $\stateset = (\statevec_1, \statevec_2, \ldots, \statevec_T)$, which are time-aligned with respect to the video frames $\frameset$.
For the $t^{\text{th}}$ video frame $\frame_t$, we have a corresponding sentence $\transtext_t$ from the text transcript and the subtask state label $\statevec_{t}$ representing the state of each subtask as a $N_{\tau}$ size vector.
Specifically, the state of the $n^{\text{th}}$ subtask at frame $t$ can have the following values: $\statescalar_t[n] \in \{$\statea, \stateb, \statec$\}$.

Our goal is to predict the subtask graph $G$ of task $\tau$ by extracting accurate subtask state from the given videos $\videoset$.
This is challenging due to two main reasons.
First, the information in the videos is noisy due to certain steps being omitted or rearranged by the video editor.
We find that the human-annotated subtask annotations violate the ground-truth subtask graph $29\%$ of the time.
Second, these subtask annotations only provide partial information about the underlying causal dependencies between the subtasks. 
We only get to observe a small amount of data where it is known with certainty that a subtask's precondition is met. 
In the majority of cases it is \emph{unknown} whether or not a subtask's preconditions are met, and worse yet, we do not have any data where we know with certainty that a subtask's preconditions are \emph{not met}.
These obstacles make this a particularly challenging learning problem and we describe how we overcome these challenges in \Cref{sec:methods}.

\section{Method} 
\label{sec:methods}
Our Multimodal Subtask Graph Generation (\ourMethodName) approach consists of two components -- subtask state prediction and graph generation.
In \Cref{subsec:method_subtask_state_prediction}, we present our subtask state prediction component which extracts missing subtask state information from video and text signals.
\Cref{subsec:method_graph_inference} describes our approach for generating subtask graphs from noisy and incomplete data from videos describing real-world activities.

\subsection{Subtask State Prediction from Noisy Annotations}
\label{subsec:method_subtask_state_prediction}

This section describes our model and training objective for predicting missing subtask state labels $\hat{\stateset}$ in the videos.
Based on the intuition that visual and textual data provide complementary information, we train a network by providing both visual and text signals to predict which of the three states ($\hat{\statescalar}_t[n] \in $ \{\emph{\statea}, \emph{\stateb}, \emph{\statec}\}) a subtask $n$ is at any given time $t$. 
These predictions are used as inputs for subtask graph generation as described in \Cref{subsec:method_graph_inference}. %

\paragraph{Architecture.}
Inspired by MERLOT~\cite{zellers-neurips21}, we jointly process visual frames and language information to model subtask states. 
Given visual information $\frameset$ and sentences from corresponding text transcript $\transtextset$, we first obtain frame embeddings $\frameset^{e}$ and sentence embeddings $\transtextset^{e}$ using a pre-trained CLIP \citep{radford-icml21} model similar to prior work \citep{buch-cvpr22,li-arxiv21,luo-arxiv21} (See~\Cref{supp_sec:subtask_state_prediction} for details about preprocessing).
We enrich the obtained visual representations with information from the text transcript through a Multi-Head Attention (MHA) mechanism~\citep{vaswani-neurips17}, followed by a residual layer as in~\Cref{eq:mha,eq:residual} (normalization and dropout omitted for brevity). 
This approach allows us to fuse misaligned visual and text information from videos.
These representations are processed by a Transformer (with a bidirectional attention mask), and we obtain $\hat{\stateset}$ as projections of the final layer Transformer representations (\Cref{eq:proj}). 
\begin{align}
    \mathbf{A} &= \mbox{MHA}(\text{Query}=\frameset^{e}, \text{Keys} = \text{Values} = \transtextset^{e}) \label{eq:mha} \\
    \hiddenset &= \text{Feed-forward}(\mathbf{A}) + \mathbf{A} \label{eq:residual} \\
    \hat{\stateset} &= \text{Feed-forward}(\text{Transformer}(\hiddenset)) \label{eq:proj}
\end{align}
Based on an intuition that sampled frame may not always include clear information (\eg, black transition frame), we predict subtask states at the beginning and the end of each subtask\footnote{based on the ground-truth subtask segmentation} (instead of predicting for every time step, which can be noisy) by appending special delimiter symbols (\tokensymbol{[start]} and \tokensymbol{[end]}, respectively) as shown in \Cref{fig:subtask_state_prediction}.
Specifically, we feed delimiter symbols to the Transformer by replacing $\hiddenset$ in \Cref{eq:proj} and estimate state predictions $\hat{\stateset}$ for a given subtask based on the final layer representations of these delimiter tokens.
We denote the final predictions based on both visual and language information as $\hat{\statevec}_t \in \mathbb{R}^{N_\tau}$.

\paragraph{Training Objective for Subtask State Prediction.} 
We incorporate ordinal regression loss~\citep{niu-cvpr16, cao-prl20} to prevent subtasks from transitioning directly from \emph{not started} to \emph{completed} without going through the intermediate \emph{in progress} state.\footnote{We observed that subtask state prediction performance dropped by $\sim$30\% when modeled as a multiclass classification problem.} 
We model our ordinal regression objective similar to~\citet{cao-prl20}, where the real line is partitioned into three subsets based on a threshold parameter $b > 0$ such that the ranges $\hat{\statescalar}_t[n] \in (-\infty, -b)$, $\hat{\statescalar}_t[n] \in (-b, b)$,  $\hat{\statescalar}_t[n] \in (b, \infty)$ respectively model the three subtask states.
As shown in \Cref{eq:masked_cross_entropy}, our objective is based on two binary classification losses corresponding to the two decision boundaries. 
$\sigma$ denotes the sigmoid function, BCE is the binary cross-entropy loss and the expectation is taken over time-steps $t$ corresponding to the special delimiter tokens and subtasks $n$ that appear in the video. 
\begin{align}
\begin{split}
    \mathcal{L}_\text{ssp} = \mathbb{E}_{t,n} 
        [ 
                &\mbox{\small BCE} ( \sigma( \hat{\statescalar}_t[n] + b), \mbb{I}[\statescalar_t[n] \neq {\small\statea}] ) \\
              + &\mbox{\small BCE} ( \sigma( \hat{\statescalar}_t[n] - b), \mbb{I}[\statescalar_t[n] =  {\small\statec}] ) 
        ]
\end{split}
\label{eq:masked_cross_entropy}
\end{align}

\subsection{Graph Generation from Subtask State Labels} %
\label{subsec:method_graph_inference}

\begin{figure*}[t]
    \centering
    \includegraphics[width=\linewidth]{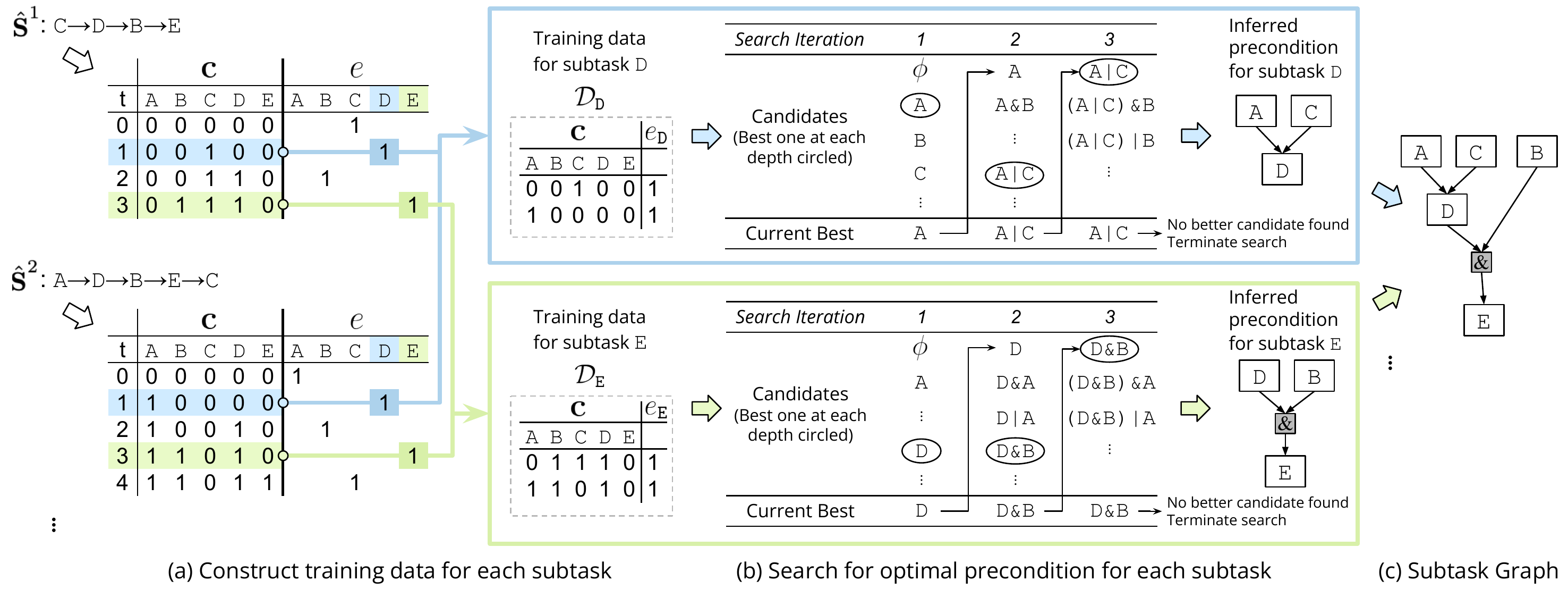}
    \caption{\textbf{Overview of Subtask Graph Generation.} 
    (a) Given the subtask state labels $\hat{\stateset}$, we extract training data $\ilpdata$ for precondition inference for each subtask.
    (b) We use a greedy algorithm to find the precondition for each subtask that maximizes the objective in \Cref{eq:jours}. At each iteration, the algorithm adds a Boolean operation and variable to construct new hypotheses. The optimal hypothesis is chosen and advanced to the next iteration. The search terminates either when there is no better solution at the current iteration or maximum number of iterations is reached.
    (c) All preconditions are consolidated into a single subtask graph.
    }
    \label{fig:ilp}
\end{figure*}

Given the noise-reduced subtask state labels $\hat{\stateset}$ we now attempt to infer the precondition for each subtask to construct the subtask graph.
\paragraph{Learning to Model Preconditions.} 
The precondition learning problem can be stated as learning a function $e_n = \pfunc(\mb{c})$ where $\mb{c} \in \{0, 1\}^{N_\tau}$ represents the completion status of each subtask (\ie, $c[i]$ denotes whether $i^\text{th}$ subtask was completed), and $e_n\in\{0, 1\}$ represents whether the precondition of $n^\text{th}$ subtask is satisfied.

The dataset $\ilpdata_n$ for inferring the precondition for $n^\text{th}$ subtask  needs to be constructed from the noise-reduced subtask state labels $\hat{\statescalar}_t[n]$.
This is non-trivial as the $\hat{\statescalar}_t[n]$ only provides partial\footnote{Previous works~\citep{hayes-icra16,huang-cvpr19, sohn-iclr20} assumed that the dataset is given and is noise-free, but these assumptions do not hold in our setting.} information about the preconditions. 
We can infer that the precondition of the $n^\text{th}$ subtask is satisfied whenever $\hat{\statescalar}_t[n]=\stateb$, since, by definition of precondition, a subtask can only be performed if its precondition is satisfied. %
However, in all other cases (\ie, $\hat{\statescalar}_t[n]\in$\{\statea, \statec\}) it is unknown whether the precondition for the subtask is met.
Based on the instances where the subtask state is $\stateb$, we construct the training dataset as shown in \Cref{eq:dataset}, where we define $\hat{c}_t[n] = \mbb{I}(\hat{\statescalar}_t[n] = \statec)$ and $\hat{e}_t[n] = \mbb{I}(\hat{\statescalar}_t[n] = \stateb)$.
\Cref{fig:ilp} (a) illustrates the data construction process.
\begin{align}
\ilpdata_n=\{(\hat{\mb{c}}_t, 1) \mid \hat{e}_t[n]=1\}
\label{eq:dataset}
\end{align}
\paragraph{Learning Objective.} 
Our learning setting presents two key challenges in graph inference.
First, the extracted data $\ilpdata$ can contain errors due to the noise in subtask state labels $\hat{\stateset}$.
Second, we only have a small amount of data (\ie, one data point per subtask per video) where a subtask's precondition is met ($e_n=1$). %
Furthermore, we have \emph{no} data with a subtask's precondition not being met ($e_n=0$).
As a result, conventional ILP algorithms that optimize $J_\text{acc}$ in \Cref{eq:ilp-org} ~\cite{muggleton-newgen91, sohn-iclr20} produce a trivial solution where there is no precondition; \ie, $J_\text{acc}$ is maximized by simply predicting $\pfunc(\mb{c})=1$ for all $\mb{c}$ since $e_n$ is always 1 in the data $\ilpdata_n$.
\begin{align}
J_{\text{acc}} 
= P(e_n = \pfunc(\mb{c})) 
&= \mathbb{E}_{(\mb{c},e_n)} {\left[\mbb{I}\left[e_n = \pfunc({\mb{c}})\right]\right]} \nonumber\\
&\simeq \mathbb{E}_{(\mb{c},e_n)\in \ilpdata_n}
\left[
    {\mbb{I}\left[e_n = \pfunc({\mb{c}})\right]}
\right]
\label{eq:ilp-org}
\end{align}
To overcome these challenges, we modify~\Cref{eq:ilp-org} and maximize the \emph{precision} as described in \Cref{eq:ilp-new}. 
\begin{align}
J_\text{prec} 
= P(e_n=1 \mid \pfunc(\mb{c})=1) 
&= \frac{
    \mathbb{E}_{(\mb{c},e_n)}\left[e_n \cdot \pfunc(\mb{c})\right]
}{
    \mathbb{E}_\mb{c} \left[\pfunc(\mb{c})\right]
} \nonumber \\
&\simeq \frac{
    \mathbb{E}_{(\mb{c},e_n)\in \ilpdata_n}\left[e_n \cdot \pfunc(\mb{c})\right]
}{
    \mathbb{E}_\mb{c} \left[\pfunc(\mb{c})\right]
} 
\label{eq:ilp-new}
\raisetag{5pt}
\end{align}
Note that $J_\text{prec} \simeq J_\text{acc}/\mathbb{E}_\mb{c} \left[\pfunc(\mb{c})\right]$ since our dataset only consists of instances with $e_n=1$. %
The denominator penalizes the precondition being overly \emph{optimistic} (\ie, predicting $\pfunc(\mb{c})=1$ more often than $\pfunc(\mb{c})=0$ will increase the denominator $\mathbb{E}_\mb{c} \left[\pfunc(\mb{c})\right]$), which helps mitigate the effect of positive label bias in the data (\ie, having no data with $e_n=0$).
Our final optimization objective is given by \Cref{eq:jours} where $\alpha$ is a regularization weight and $C(\pfunc)$ measures the complexity of the precondition $\pfunc$ in terms of the number of Boolean operations.
The regularization term handles the data noise since the noise often adds extra dependency between subtasks and results in overly complicated (\ie, large number of Boolean operations) precondition.
See~\supplementary{\Cref{supp_subsec:ilp_ours}}~for more details.
\begin{equation}
\max_{\pfunc}J_\text{ours}(\pfunc) = \max_{\pfunc}\ J_\text{prec}(\pfunc) - \alpha C(\pfunc)
\label{eq:jours}
\end{equation}
\paragraph{Optimization.}
Since \Cref{eq:jours} is an NP-hard optimization problem, we consider a greedy search algorithm to find a good precondition $\pfunc$. 
Starting from the null precondition, at each iteration of the search, we construct candidate preconditions by adding a Boolean operation (\eg, $\&$ and $\mid$) and variable (\eg, \subtask{A}, \subtask{B}, etc) to the best precondition identified in the previous iteration.
We choose the candidate precondition that maximizes \Cref{eq:jours} and continue to the next iteration.
The search terminates either when a maximum number of iterations is reached or no better solution is found in the current iteration.
See \Cref{fig:ilp} (b) for an illustration of the search algorithm.

\section{Experiments}
\label{sec:experiments}

We first evaluate the graph generation performance of our method in \Cref{subsec:experiment_graph_generation}. 
Then, we apply our approach to a downstream task of next subtask prediction in \Cref{subsec:experiment_next_step_prediction}.

\subsection{Graph Generation}
\label{subsec:experiment_graph_generation}

In this section, we qualitatively and quantitatively compare our method against baselines on the subtask graph generation task.
We first introduce the dataset and the baselines we use and then discuss the main results.

\paragraph{Dataset.} %
Since there are no existing datasets that consist of videos with subtask graph annotations, we constructed a new dataset based on the ProceL dataset~\citep{elhamifar-iccv19}.
ProceL includes a variety of subtasks with dense labels from real-world instructional videos.
We asked three human annotators to manually construct a graph for each task and aggregated these annotations by retaining the high-confidence preconditions (See~\Cref{supp_subsec:ablation_study_subtask_state_prediction} for more details).
We use all the videos from twelve primary tasks in ProceL accompanied by spoken instructions generated using Automatic Speech Recognition (ASR).
Each task has 54.5 videos and 13.1 subtasks on average, and each video has subtask label and timing annotations (start and end times).

\paragraph{Baselines.} %
We compare our approach against the following baselines. 
\begin{itemize}
\item \textbf{MSGI}~\cite{sohn-iclr20}: An inductive logic programming (ILP) method that maximizes the objective in \Cref{eq:ilp-org}. %
\item \textbf{\msgip}: A variant of MSGI which assumes that the precondition is not met until the subtask is performed (\ie, $e_t[n]=0$ if $\hat{\statescalar}_t[n]=\statea$) similar to prior work \citep{hayes-icra16, xu-icra18}). 
We demonstrate that this method, though an improvement over MSGI, yields worse performance than \ourMethodName{} due to its strong assumption.
We instead assume that it is unknown whether precondition is met in this case (\ie, $e_t[n]$ is unknown for $\hat{\statescalar}_t[n]\neq\stateb$).

\item \textbf{\ourGraph} (\ourGraphFull): Generate graph (\Cref{subsec:method_graph_inference}) from human-annotated subtask state labels $\stateset$. In other words, this is our method without the multimodal subtask state inference module in \Cref{subsec:method_subtask_state_prediction}.
\item \textbf{ProScript}~\cite{sakaguchi-emnlpf21}: A T5 model~\cite{raffel-jmlr20} fine-tuned to predict a partially-ordered script from a given scenario description (\eg, baking a cake) and a set of unordered subtask descriptions. 
Note that this is a text-only baseline trained on script data from \citet{sakaguchi-emnlpf21} and does not use the ProceL video/text data.
\end{itemize}
We use T5-Large~\cite{raffel-jmlr20} as the transformer in our state prediction module.
During training we randomly omit video frames corresponding to 25\% of subtasks for data augmentation.
See~\supplementary{\Cref{supp_subsec:ablation_study_subtask_state_prediction}} for further details and ablations regarding choice of model and architecture.

\paragraph{Metrics.}
We consider the following metrics to measure the quality of predicted subtask graph $G=(\pfuncarg{1}, \ldots, \pfuncarg{N_\tau})$.

\begin{itemize}
\item \textit{Accuracy} measures how often the output of the predicted and the ground-truth preconditions agree~\cite{sohn-iclr20}, where 
$\pfunc^{*}$ is the ground-truth precondition of the $n^\text{th}$ subtask. 
\begin{align}
    \text{Accuracy}=\frac{1}{N_{\tau}}\sum_{n=1}^{N_{\tau}} \mathbb{E}_\mb{c} \left[
    \ind\left[
    \pfunc(\mb{c})=\pfunc^{*}(\mb{c}) 
    \right]
    \right]
    \label{eq:accuracy}
\end{align}
\item \emph{\ourGraphMetricFull}~(\ourGraphMetric): We define a strict partial order relation $R_G$ imposed by graph $G$ on subtasks $a, b \in S$ as: $(a,b) \in R_{G}$ iff $a$ is an ancestor of $b$ in graph $G$ (\ie, it can be written as a binary relation with matrix notation: $R_{G}[a][b] \triangleq \mathbb{I}$[$a$ is an ancestor of $b$ in $G$]).
\ourGraphMetric~is defined as in \Cref{eq:graphmetric}, where $G^{*}$ denotes the ground-truth graph.
\begin{align}
\label{eq:graphmetric}
\text{\ourGraphMetric}=\frac{1}{N_{\tau}(N_{\tau}-1)}  \sum_{a\neq b}\mathbb{I}\left[R_{G}[a][b]=R_{G^{*}}[a][b]\right]
\raisetag{10pt}
\end{align}

\item \emph{\ourDataMetricFull} measures the extent to which subtask trajectories in the dataset are compatible with causality constraints imposed by the predicted graph.
For instance, given a subtask sequence $x_1,\ldots,x_n$, if $x_j$ is an ancestor of $x_i$ in the graph (\ie, $(x_j, x_i) \in R_{G}$) for $i < j$, this might suggest that subtasks $x_i$ and $x_j$ violate the causality constraint imposed by the graph.
However, this is not always the case as a subtask can still occur \emph{before} an \emph{ancestor} subtask as long as it's precondition is satisfied (\ie, $\pfuncarg{x_i}(x_1,\ldots,x_{j-1})=1$).\footnote{In an abuse of notation, we use $\pfuncarg{x_i}(x_1,\ldots,x_{j-1})$ to represent whether the precondition of $x_i$ is satisfied given that subtasks $x_1, \ldots,x_{j-1}$ were completed.}
We thus define the \ourDataMetric~of the given subtask sequence as in \Cref{eq:compatibility}.\footnote{We assume the product taken over a null set to be 1.}
The metric is (macro) averaged across all dataset instances.
\begin{align}
\text{\ourDataMetric} = \frac{1}{n} \sum_{j=1}^n \prod_{\substack{i < j \\ {(x_j, x_i) \in R_{G}}}} \pfuncarg{x_{i}}(x_1,\ldots,x_{j-1})
\label{eq:compatibility}
\raisetag{13pt}
\end{align}

\end{itemize}

\begin{table}[t]
    \setlength{\tabcolsep}{4pt}
    \centering
    \small
    \caption{\textbf{Graph generation results in ProceL~\cite{elhamifar-iccv19}.}
    We report the average percentage (\%) across all tasks (See~\supplementary{\Cref{supp_subsec:task_level_graph_generation_result}} for task-level performance).
    }
    \begin{tabular}{l|ccc}
        \toprule  
        Method & Accuracy~$\uparrow$ & \ourGraphMetric~$\uparrow$ & \ourDataMetric~$\uparrow$ \\ \midrule
        ProScript & 57.50 & 62.34 & 37.12 \\ 
        MSGI & 54.23 & 64.62 & N/A \\
        \msgip & 73.59 & 72.39 & 76.08 \\
        \ourGraph & 81.62 & 88.35 & 97.86 \\ 
        \textbf{\ourMethodName~(Ours)}  & \textbf{83.16} & \textbf{89.91} & \textbf{98.30} \\ 
        \bottomrule
    \end{tabular}
    \label{tab:graph_evaluation}
\end{table}
\begin{figure*}[t]
    \centering
    \includegraphics[width=0.95\linewidth]{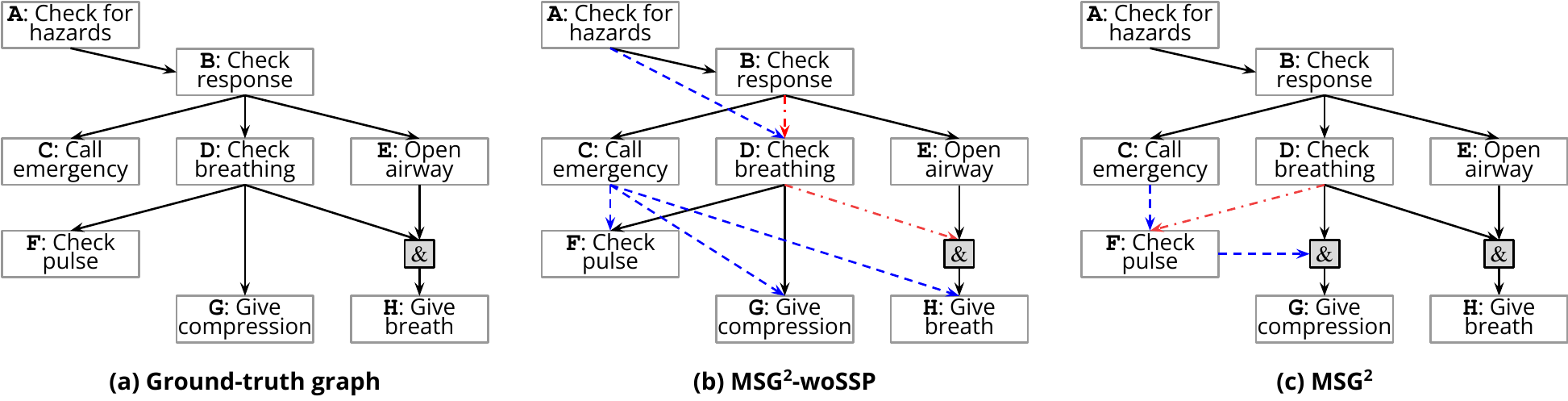}
    \caption{\textbf{Illustration of subtask graphs generated by each method for \task{\maintask} task}. %
    For the predicted graphs ((b)~\ourGraphFull~and (c)~\ourMethodName), {\color{blue}redundant edges} (\ie, false positive) are colored in {\color{blue}blue} and {\color{red}missing edges} (\ie, false negative) are colored in {\color{red}red}.
    Please check~\Cref{supp_fig:generated_graph_examples} for more examples.
    }
    \label{fig:generated_graph}
\end{figure*}

\paragraph{Results.}
\Cref{tab:graph_evaluation}~summarizes the performance of different graph generation methods. 
First, the complexity regularization term~(\Cref{eq:ilp-new}) helps our method (\ourMethodName) be more robust to incomplete and noisy data compared to the MSGI and \msgip baselines.
Intuitively, \msgip assumes that the input data contains no noise and attempts to infer a binary decision tree that perfectly discriminates between eligible and ineligible samples. 
When the data is noisy, we observe that such strictness results in an overly complicated precondition (\ie, overfits to the noise), resulting in inaccurate graphs.

On the other hand, MSGI simply ignores all the data points when it is unknown whether precondition is met.
This results in MSGI predicting \emph{null} preconditions (\ie, precondition is always satisfied) for all subtasks.
We thus report \emph{\ourDataMetricFull}~as N/A in \Cref{tab:graph_evaluation} for MSGI. 
Note that our \ourMethodName{} circumvents this degeneration problem by optimizing the precision (as opposed to the accuracy, as in MSGI) and regularizing the complexity of the inferred graph (see \Cref{eq:jours}).

When using subtask states predicted by our state prediction module (\ourMethodName) instead of human annotations (\ourGraph), we observe consistent improvements for all metrics, which shows the benefit of predicting missing subtasks by exploiting multimodal vision-text data. 
See~\supplementary{\Cref{supp_fig:completion_prediction_examples}}~for examples of subtask state prediction.
Our method also significantly outperforms proScript~\cite{sakaguchi-emnlpf21}, which relies on a web-scale pretrained text model~\cite{raffel-jmlr20}.

\paragraph{Qualitative Evaluation.} %
\Cref{fig:generated_graph}~shows predicted graphs for the \task{\maintask}~task.
The~\ourMethodName~graph (Figure~\ref{fig:generated_graph}.(c)) is closer to the ground-truth graph (Figure~\ref{fig:generated_graph}.(a)) compared to our model without subtask state prediction (Figure~\ref{fig:generated_graph}.(b)).
This is because the predicted subtask states provide additional clues about subtask completion.
For instance, consider the subtask \subtask{D:check breathing}, which is a prerequisite for subtasks \subtask{F}, \subtask{G} and \subtask{H} in the ground truth graph.
Since \subtask{D} is sparsely annotated in the data (does not appear in 29\% of the sequences), the baseline model assumes that the presence of other subtasks (\eg, \subtask{C}) explains subtasks \subtask{F}, \subtask{G} and \subtask{H} (the redundant outgoing arrows from \subtask{C}).
By recovering some of the missing annotations for \subtask{D} based on visual and text signals (\eg, it is clearly visible that \subtask{check breathing} was performed), our method resolves some of these cases (\eg, \subtask{C} is no longer a precondition for \subtask{H}).
However, the recovery is not perfect, and our model still mistakenly assumes that \subtask{C} is a precondition for \subtask{F}. 
Further improvements in multimodal subtask state modeling can help address these errors.

\subsection{Next Subtask Prediction}
\label{subsec:experiment_next_step_prediction}

Inspired by~\citet{epstein-iccv21}, which performed the downstream task of predicting the next \emph{subtask}, we consider the next subtask prediction based on the inferred subtask graph. %
Predicting the next subtask can help humans and agents break down a complex task into more manageable subtasks and focus on subtask that are currently relevant.
It further helps complete tasks efficiently by minimizing trial and error, and identify/address potential issues that may arise during the task. 
In contrast to predicting the next \emph{frame} in time~\cite{damen-eccv18,kuehne-cvpr14}, next \emph{subtask} prediction requires reasoning about events at a higher level of temporal abstraction. %
\paragraph{Dataset.} %
We use the CrossTask dataset, which consists of 18 primary tasks with 7.4 subtasks and 140.6 videos per task on average.\footnote{We found CrossTask to be more interesting and challenging than other datasets in the literature as it includes more variation in subtasks.}
We also adopt all 12 tasks from the ProceL dataset which includes 13.2 subtasks with 54.5 videos per task on average.  
For both ProceL~\cite{elhamifar-iccv19} and CrossTask~\cite{zhukov-cvpr19}, we convert all videos to 30 fps, obtain verb phrases and extract CLIP features following~\Cref{subsec:experiment_graph_generation}.
For each dataset we take 15\% of the videos of each task as the test set.
Please see~\supplementary{\Cref{supp_sec:next_step_pred}}~for the details.
\paragraph{Baselines.} %

We first choose two open-source end-to-end video-based Transformer models (STAM \cite{sharir-arxiv21}, ViViT \cite{arnab-iccv21}) for comparison.
These models are based on the Vision Transformer~\cite{dosovitskiy-iclr21}, which slices each frame into $16\times16$ patches and treats each patch as an input to the transformer.
To keep the input the same for all models used in this experiment, we replace the patch embedding from each model implementation with the CLIP embedding used in our model.
Specifically, we replace the patch embedding with $\hiddenset^{i}$~in~\Cref{eq:residual} for each model.
We train the methods to predict the next subtask from the history of previous subtasks using a multi-class subtask classification loss, where a linear projection of the final transformer hidden state is used as the subtask prediction.

In addition to these end-to-end neural baselines, we also tested the graph-based methods introduced in~\Cref{tab:graph_evaluation}.
For all graph-based methods, we first generated the graph from the train split.
We then use the generated subtask graph to predict whether the precondition is satisfied or not from the subtask completion in test split.
Specifically, we used the GRProp policy~\cite{sohn-neurips18}, which is a simple GNN policy that takes the subtask graph as input and chooses the best next subtask to execute.
It aims to maximize the total reward, when each subtask is assigned with a scalar-valued reward.
At every time step, we assigned higher reward to each subtask if its precondition was satisfied more \emph{recently}: \ie, $r_n \propto \lambda^{{\Delta t}_n}$, where $r_n$ is the reward assigned to $n^\text{th}$ subtask, ${\Delta t}_n$ is the time steps elapsed since the precondition $\pfunc$ for has been satisfied and $0<\lambda < 1$ is the discount factor.
Please see~\supplementary{\Cref{supp_subsec:next_step_pred_details}} for more details.

\paragraph{Metric.} %
For each test video, we measure the accuracy of predicting each subtask correctly, given previous subtasks and corresponding ASR sentences.

\begin{table}[t]
    \centering
    \small
    \setlength{\tabcolsep}{4pt}
    \caption{\textbf{Next Subtask Prediction Task.} We perform the next subtask prediction on ProceL~\cite{elhamifar-iccv19}~and CrossTask dataset~\cite{zhukov-cvpr19}~and measure the accuracy (\%). 
    We report the average among all tasks, and task-level accuracy can be found in~\supplementary{\Cref{supp_subsec:task_level_next_step_pred_result}}.
    }
    \begin{tabular}{l|cc}
    \toprule  
    Model & ProceL & CrossTask \\ \midrule
    STAM~\citep{sharir-arxiv21} & 29.86 & 40.17 \\
    ViViT~\citep{arnab-iccv21} & 26.98 & 41.96 \\\midrule
        proScript~\cite{sakaguchi-emnlpf21} & 18.86 & 36.78 \\ 
    MSGI~\citep{sohn-iclr20} & 17.42 & 32.31 \\
    \msgip & 26.54 & 32.72 \\
    \ourGraph & 48.39 & 53.39 \\ 
    \textbf{\ourMethodName~(Ours)} & \textbf{55.38} & \textbf{54.42} \\
    \bottomrule
    \end{tabular}
    \label{tab:next_task_pred}
\end{table}

\paragraph{Results.}
\Cref{tab:next_task_pred} shows the next subtask prediction performance.
Compared to the end-to-end neural baselines, \ourMethodName~achieves 85\% and 30\% higher prediction performance in ProceL~\cite{elhamifar-iccv19} and CrossTask~\cite{zhukov-cvpr19} datasets, respectively.
In addition, we observe that end-to-end neural baselines are competitive with some of the graph generation baselines.
This indicates that the higher quality of graphs predicted by our approach is responsible for the significant performance improvement over all baselines.

\section{Related Work}
\label{sec:related}

\subsection{Learning Compositional Tasks using Graphs} %
\label{subsec:related_task_graph}
Prior works have tackled compositional tasks by modeling the task structure using a manually defined symbolic representation. 
Classical planning works learn STRIPS~\cite{fikes-ai71, frank-c03} domain specifications (action schemas) from given trajectories (action traces)~\cite{mehta-neurips11,suarez-aaaiw20,walsh-aaai08,zhuo-ai10}.
Reinforcement learning approaches model task structure in the form of graphs such as hierarchical task networks~\cite{hayes-icra16,huang-cvpr19,xu-icra18} and subtask graphs~\cite{liu-aaai22,sohn-neurips18,sohn-iclr20} to achieve compositional task generalization. 
However, they assume that there is no noise in the data and high-level subtask information (\eg, whether each subtask is eligible and completed) is available to the agent.
Thus, existing methods cannot be directly applied to videos of real-world activities due to (1) missing high-level subtask information and (2) data noise (\eg, rearranged, skipped, or partially removed subtasks).
Instead, our model learns to predict subtask progress from the video and adopts a novel inductive logic programming algorithm that can robustly generate the subtask graph from noisy data.

\subsection{Video Representation Learning} %
\label{subsec:related_vision_language}
There has been limited progress in video representation learning~\cite{simonyan-neurips14,tran-iccv15} compared to its image counterpart~\cite{he-cvpr16,he-iccv17,huang-cvpr17}, due in part to the considerable human effort in labeling innately higher-dimensional data~\cite{miech-iccv19}.
Following the success of the Transformer \citep{vaswani-neurips17} architecture in Natural Language Processing~\cite{devlin-naacl19,radford-blog19}, it has been subsequently used for visual representation learning~\cite{li-arxiv19,radford-icml21,ramesh-icml21}. 
More recent efforts have been directed towards learning from multimodal video and text data \cite{miech-iccv19,miech-cvpr20,sun-iccv19,zellers-neurips21}.
One of the key issues in learning from video and paired text narrations is handling the misalignment between video frames and text transcripts~\cite{miech-cvpr20,shen-cvpr21}.
In addition, extracting a holistic understanding of video content is challenging, so most prior works focus on short clip-level representations \cite{miech-iccv19,miech-cvpr20,sun-iccv19}.
Our work attempts to extract a holistic understanding of a task described in instructional videos by learning to jointly model visual and language information.
Unlike prior works that study the sequence of events appearing in individual instructional videos \cite{elhamifar-iccv19,zhukov-cvpr19,tang-cvpr19}, we attempt to consolidate information from multiple videos of a task to obtain a robust understanding of the task structure.

\section{Conclusion}
\label{sec:conclusion}

We present~\ourMethodName, a method for generating subtask graphs from instructional videos on the web that exhibit subtask rearrangement, skipping, and other annotation errors.
We first predict the state of each subtask from individual videos and use it as input to obtain better subtask graphs. 
We additionally propose a novel complexity-constrained ILP method to deal with noisy and incomplete data.
We demonstrate that our method produces more accurate subtask graphs than baselines.
In addition, our method achieves better performance than recent video transformer approaches in predicting the next subtask, demonstrating the usefulness of our subtask graph. 

We plan to improve \ourMethodName~in several directions:
1) We used the text generated from ASR and the frames from videos as the main source of learning the task.
Employing other modalities, such as audio or motion-level features, may help to improve the quality of prediction, thereby generating more accurate subtask graphs.
2) Our model generates graphs in two stages, subtask state prediction and graph generation, so information can be lost in the middle.
For instance, details about the video could be lost at the end of the subtask state prediction stage.
A potential approach is to design an end-to-end network that directly computes a graph from video features.
3) In this work, we did not consider task transfer.
\ourMethodName~learns about each task individually.
Transferring knowledge across different tasks could be a promising direction.

\subsection*{Acknowledgements}
We thank Jae-Won Chung, Junhyug Noh and Dongsub Shim for constructive feedback on the manuscript.
This work was supported in part by grants from LG AI Research and NSF CAREER IIS-1453651.

{\small
\bibliographystyle{abbrvnat}
\bibliography{arxiv_techreport2023}
}

\clearpage
\appendix
\onecolumn

\renewcommand{\thetable}{\Alph{table}}
\renewcommand{\thefigure}{\Alph{figure}}
\renewcommand{\thesection}{\Alph{section}}
\renewcommand{\theequation}{\Alph{equation}}
\renewcommand{\thesubsection}{\thesection.\alph{subsection}}
\setcounter{figure}{0}    
\setcounter{table}{0}    
\setcounter{equation}{0}

\section{Subtask State Prediction}
\label{supp_sec:subtask_state_prediction}

\subsection{Training Details}
\label{supp_subsec:training_details_subtask_state_prediction}

\paragraph{Preprocessing.} We first convert all videos in ProceL~\cite{elhamifar-iccv19} to 30 fps.
Then, we extract the verb phrases of \textit{verb+(prt)+dobj +(prep+pobj)}\footnote{parenthesis denotes optional component, prt: particle, dobj: direct object, prep: preposition, pobj: preposition object.} from Automated Speech Recognition (ASR) output, using the implementation in SOPL~\cite{shen-cvpr21}.
For both vision and text data, we extract the frame and verb phrases features from the `ViT-B/32' variant of CLIP~\cite{radford-icml21}.
We extract the features with each subtask's start and end position and load the CLIP features, instead of video frames, for faster data reading.
We choose the first label of each subtask in a video and convert the temporal segment labels to \statea,~\stateb, and~\statec~for the existing subtasks. %

\paragraph{Training.} We set the hidden dimension of each feedforward dimension in 16-head modality fusion attention to be the same as the CLIP representation embedding size of 512 and use a single fully-connected layer when projecting to the status prediction.
During training, we randomly drop up to 25\% of subtasks from each task sequence and then subsample at least three frames from each subtask region, but no more than 190 frames in total. 
On the other hand, for testing, we did not drop any subtask and sample 3 frames per subtask in an equidistance manner (no randomness).
For all the language features, because the length of the ASR varies from video to video, we use three consecutive sentences per frame while setting the center of the sentence closest to the selected frame, inspired by~\citet{miech-cvpr20}.
We train all models with a learning rate of 3e-4 with the Adam~\cite{kingma-iclr15} optimizer with cosine scheduling, following BERT~\cite{devlin-naacl19}.
We set the batch size as 32 and trained each model for 600 epochs, with 100 steps of warm-up.
Each model is trained on an 18.04 LTS Ubuntu machine with a single NVIDIA A100 GPU on CUDA 11.3, cuDNN 8.2, and PyTorch 1.11.

\subsection{Ablation Study}
\label{supp_subsec:ablation_study_subtask_state_prediction}
\begin{table}[ht]
    \setlength{\tabcolsep}{3.5pt}
    \centering
    \footnotesize
    \caption{\textbf{Ablation Result on Subtask State Prediction in ProceL~\cite{elhamifar-iccv19}.} We denote video-input only case as ``VisionOnly'', video with the narration text data, but with skip connection as ``Vision + ASR'', and our subtask state prediction model as ``\textbf{Ours}'' and measure the performance of completion prediction. We denote the pretrained transformer model as ``ViT'' and ``VisualBERT'', following the pretrained model names~\cite{dosovitskiy-iclr21, li-arxiv19}. Task label indexes are (a) Assemble Clarinet (b) Change Tire (c) Perform CPR (d) Setup Chromecast (e) Change Toilet Seat (f) Make Peanut Butter and Jelly Sandwich (g) Jump Car (h) Tie Tie (i) Change iPhone Battery (j) Make Coffee, (k) Repot Plant and (l) Make Salmon Sandwich, respectively.}
    \begin{tabular}{l|cccccccccccc|c}
    \toprule  
    Subtask State Prediction Module & (a) & (b) & (c) & (d) & (e) & (f) & (g) & (h) & (i)& (j) & (k) & (l) & Avg\\
    \midrule
    VisionOnly & 72.41 & 74.91 & 80.21 & 93.68 & 72.94 & 91.11 & 85.58 & 93.65 & 89.52 & 88.93 & 79.45 & 70.90 & 82.77 \\ 
    Vision + ASR & 71.83 & 74.76 & 83.12 & 89.42 & 69.05 & 91.07 & 83.80 & 90.08 & 86.36 & 87.78 & 78.16 & 72.64 & 81.51 \\ 
    Ours (from scratch) & 63.56 & 68.93 & 74.73 & 90.36 & 66.61 & 79.02 & 71.54 & 87.76 & 69.54 & 80.47 & 72.42 & 65.97 & 74.24 \\
    Ours (w/o~\stateb~state) & 70.67 & 74.42 & 82.72 & 93.28 & 72.41 & 89.95 & 85.56 & 93.94 & 88.01 & 88.85 & 79.59 & 74.63 & 82.84 \\
    \textbf{Ours} & 71.25 & 74.93 & 83.51 & 94.56 & 73.42 & 89.76 & 85.94 & 94.71 & 91.14 & 89.54 & 79.44 & 75.58 & \textbf{83.65} \\\midrule
    VisualBERT~\cite{li-arxiv19} & 69.09 & 73.84 & 79.63 & 69.59 & 58.45 & 91.68 & 84.01 & 90.40 & 83.18 & 83.83 & 62.07 & 67.73 & 76.13 \\ 
    VisualBERT$\dagger$ & 59.01 & 59.99 & 72.98 & 71.07 & 64.66 & 78.93 & 77.69 & 72.33 & 75.52 & 75.33 & 63.43 & 68.10 & 69.92 \\ 
    ViT~\cite{dosovitskiy-iclr21} & 58.69 & 59.92 & 74.11 & 68.84 & 59.69 & 79.09 & 78.29 & 70.39 & 74.90 & 78.17 & 70.03 & 67.45 & 69.96 \\ 
    \bottomrule
    \end{tabular}
    \label{supptab:frame_evaluation}
\end{table}
Since a subtask sequence $\stateset$ in video stores the start and end frame numbers, we can directly compare the completion prediction result of all subtasks by checking $\mbb{I}[\hat{\statescalar}_t[n] \geq 1]$.
However, because the label for $\stateset$ only covers the labeled part of the sequence, we first split 15\% of the data as the validation set and hand-annotated the subtask state of all subtask labels for the videos in the validation set.
For both the ground-truth subtask graph and the subtask state labels, we asked three people to manually annotate after watching all videos in the task. We choose the majority answer among three as the label for the subtask state labels, and we iterate multiple rounds of ground-truth subtask graph labeling until the label converges among three people.
We performed an ablation study with the VisionOnly (replacing $\hiddenset$ defined in~\Cref{eq:residual} with $\frameset^{e}$), our model without having first binary cross entropy loss in~\Cref{eq:masked_cross_entropy} (denoted as~w/o~\stateb~state), as well as our model with the skip connection (adding $+ \frameset^{e}$ to the right-hand side of~\Cref{eq:mha}, following Transformer~\cite{vaswani-neurips17}. We denote this as `Vision+ASR') by measuring the binary completion prediction accuracy per task with the hand-annotated labels. 
In addition to this, we performed additional experiments with the pretrained ViT~\cite{dosovitskiy-iclr21} and VisualBERT~\cite{li-arxiv19} models from Huggingface~\cite{wolf-arxiv19}, replacing the T5~\cite{raffel-jmlr20} model. Specifically, we use `google/vit-large-patch16-224', `uclanlp/visualbert-vqa-coco-pre', and `google/t5-v1\_1-large' pretrained weights for this ablation.
We also tried a variation of VisualBERT (indicated as VisualBERT$^{\dagger}$) where we directly feed the frames $\frameset^{e}$ and sentences $\transtextset^{e}$, instead of $\hiddenset$ in~\Cref{eq:residual}, as input. We set $b$ in~\Cref{eq:masked_cross_entropy} as 1 for all of the models.

The results are shown in~\Cref{supptab:frame_evaluation}. %
First of all, we can see that our model with the skip connection could lead the model to be overfitted to the train set, which is the reason behind our decision not to add a skip connection to the~\ourMethodName~model.
Also, we found inferior performance when we train without ASR or~\stateb~state information, so we perform subtask state prediction with both vision and language modality with~\stateb~for the rest of the paper. %
We additionally found that the model predicts~\statec~for the unlabeled tasks more clearly one subtask after the original prediction timing.
We conjecture that the end-time of a subtask, which is annotated by a human annotator, is often noisy and annotated to a slightly earlier time step where subtask is still \stateb.
Thus, we grab the states from the next subtask timing and use \statec~if $\hat{\statescalar}[n] \geq 0$ instead in graph generation.
We also trained our model from scratch instead of finetuning a pretrained T5 encoder-based model. We believe the performance gap between `Ours' and `Ours (from scratch)' shows the effectiveness of finetuning from the pretrained weights. %
In addition to this, we tested with other pretrained Transformers and found that the pretrained T5 encoder-based model performs best among all the transformer models.
Interestingly, ViT and VisualBERT were worse than T5, which seems to indicate that language priors are more useful for modeling subtask progression.

\subsection{Visualization of Subtask Completion}
\label{supp_subsec:visualization_of_subtask_completion}

\begin{figure}[ht]
    \centering
    \begin{subfigure}{\linewidth}
            \centering
            \includegraphics[width=0.98\linewidth]{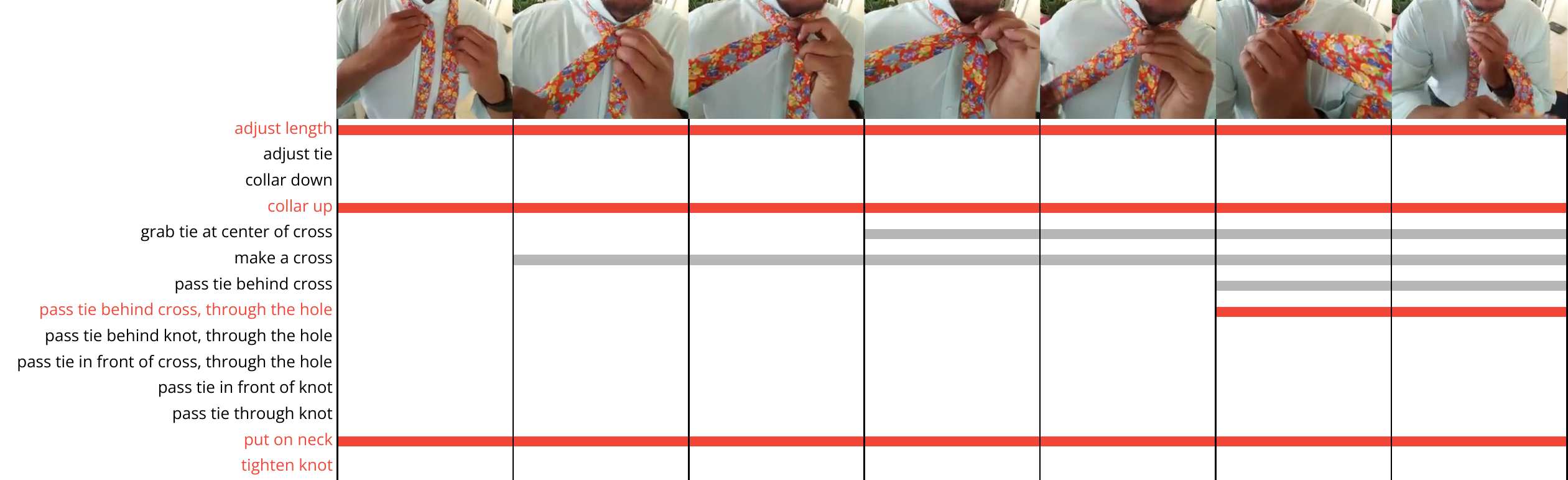}
            \caption{Tie Tie}
    \end{subfigure}
    
    \begin{subfigure}{\linewidth}
            \centering
            \includegraphics[width=0.98\linewidth]{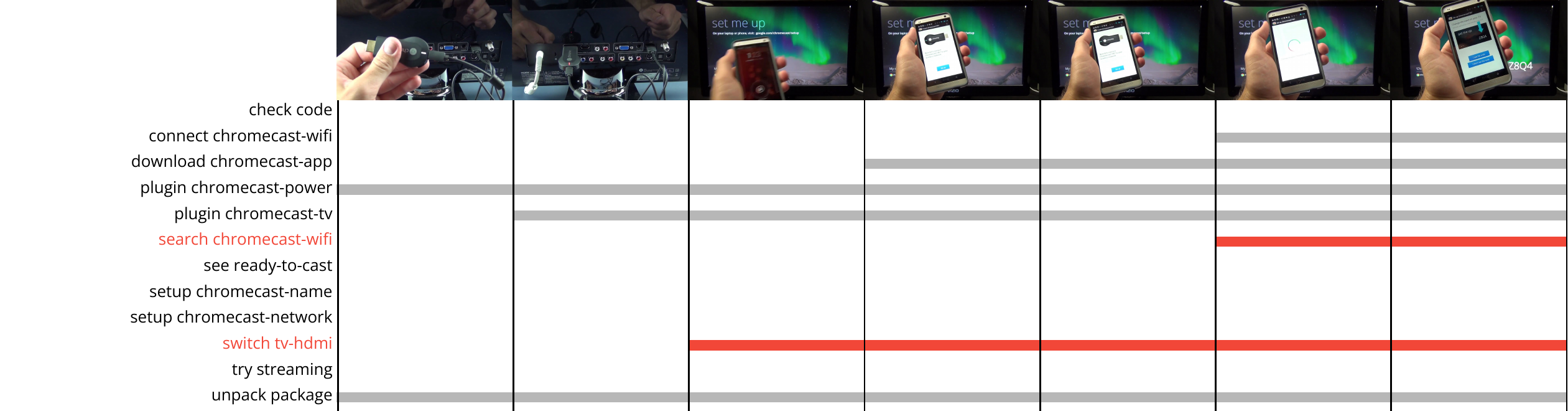}
            \caption{Setup Chromecast}
    \end{subfigure}
    \caption{\textbf{Completion Prediction Examples.} We plot predicted subtask completions from (a) Tie Tie and (b) Setup Chromecast. The missing labels in the original dataset are colored red, and each colored row represents the completion of a subtask at the matched frame on top.
    The presence of a horizontal bar at a particular time-step indicates the subtask is in \statec~state at the time-step.
    Bars in red show subtask completion states that were inferred by our model but were missing in the dataset.
    }
    \label{supp_fig:completion_prediction_examples}
\end{figure}

We present predicted subtask completion in~\Cref{supp_fig:completion_prediction_examples}. 
Our model predicts~\statec~states from missing subtasks (labeled in {\color{red}red}).
Such predicted subtask states help generate better graphs.

\section{Graph Generation}
\label{supp_sec:ilp_detail}

\subsection{Background: Subtask Graph Inference using Layer-wise Inductive Logic Programming}
\label{supp_subsec:ilp_prelim}

For a task $\tau$ that consists of $N_\tau$ subtasks, we define the \emph{completion} vector $\smash{\mb{c}\in\{0, 1\}^{N_\tau} }$ and \emph{eligibility} vector $\smash{\mb{e}\in\{0, 1\}^{N_\tau} }$ where $c_n$ indicates if the $n$-th subtask has completed and $e_n$ indicates if the $n$-th subtask is eligible (\ie, its precondition is satisfied).
Given $\ilpdata=\{(\mb{c}^j, \mb{e}^j)\}_{j=1}^{|\ilpdata|}$ as training data,~\citet{sohn-iclr20}~proposed an Inductive Logic Programming (ILP) algorithm which finds the subtask graph $G$ that maximizes the binary classification accuracy (\Cref{suppeq:ilp-obj}):
\begin{align}
    \hat{G}
    &=\argmax_{G} P(\pfuncarg{G} (\mb{c}) = \mb{e} )\label{suppeq:ilp-obj}\\
    &=\argmax_{G} \sum_{j=1}^{|\ilpdata|}{ \mbb{I}[\mb{e}^{j} = \pfuncarg{G}(\mb{c}^{j})] }\\
    &= 
    \left(
        \argmax_{G_1} \sum_{j=1}^{|\ilpdata|}{ \mbb{I}\left[e^{j}_1 = \pfuncarg{1}(\mb{c}^{j})\right] },
        \ldots,
        \argmax_{G_{N_\tau}} \sum_{j=1}^{|\ilpdata|}{ \mbb{I}\left[e^{j}_{N_\tau} = \pfuncarg{N_\tau}(\mb{c}^{j})\right] }
    \right),\label{suppeq:ilp-indiv}
\end{align}
where $\mbb{I}[\cdot]$ is the element-wise indicator function, $\pfuncarg{G}: \mb{c}\mapsto \mb{e}$ is the precondition function defined by the subtask graph $G$, which predicts whether subtasks are eligible (\ie, the precondition is satisfied) from the subtask completion vector $\mb{c}$, and $\pfunc: \mb{c}\mapsto e_n$ is the precondition function of $n$-th subtask defined by the precondition $G_n$.

\begin{figure*}[t]
    \centering
    \includegraphics[draft=false,width=0.99\linewidth]{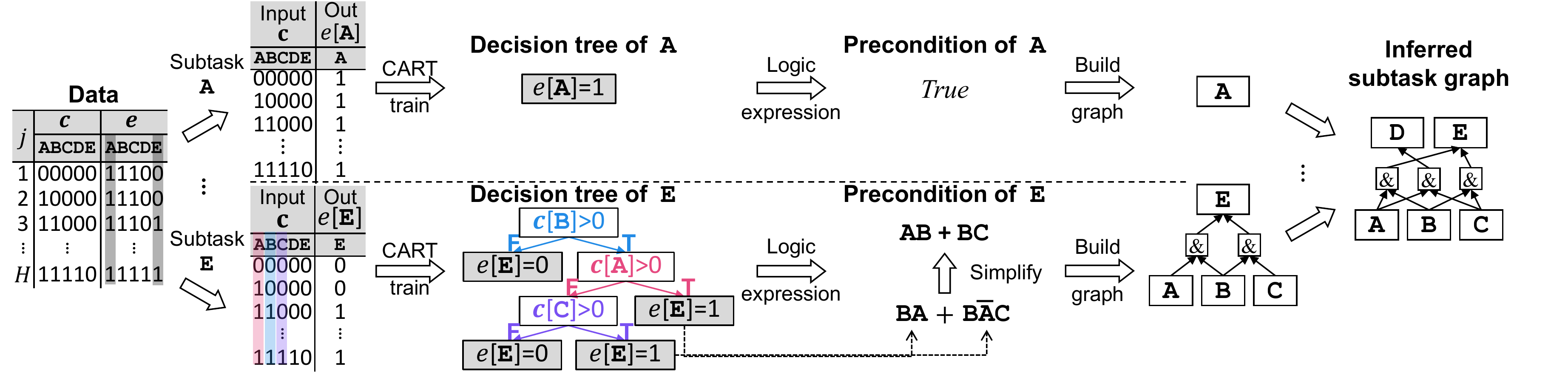}
    \caption{
        The inductive logic programming (ILP) module takes all the completion and eligibility vectors ($\{(\mb{c}^j, \mb{e}^j)\}_{j=1}^{|\ilpdata|}$) as input, and builds a binary decision tree to infer the precondition $G$. 
        For example, in case of subtask \subtask{E} (the bottom row in the Figure), it takes ($\{(\mb{c}^j, e^j[\text{\subtask{E}}])\}_{j=1}^{|\ilpdata|}$) as (input, output) pairs of training dataset, and constructs a decision tree by choosing a variable (\ie, component of completion vector $\mb{c}$, which corresponds to each subtask) at each step that best splits the true (\ie, $e[\text{\subtask{E}}]=1$) and false (\ie, $e[\text{\subtask{E}}]=0$)-labeled data samples. 
        Then, the decision tree is represented as a logic expression (\ie, precondition), simplified, transformed to a subtask graph form, and merged together to form a subtask graph.
        The figure was adopted from~\citet{sohn-iclr20}~and modified to match our notation style. 
    }
    \label{supp_fig:ilp}
\end{figure*}

\Cref{supp_fig:ilp} illustrates the detailed process of subtask graph inference in~\citet{sohn-iclr20} from the completion and eligibility data.
The precondition function $\pfunc$ is modeled as a binary decision tree where each branching node chooses the best subtask (\eg, subtask B in the first branching in the bottom row of~\Cref{supp_fig:ilp}) to predict whether the $n$-th subtask is eligible or not based on Gini impurity~\cite{breiman-routledge84}.
Each binary decision tree constructed in this manner is then converted into a logical expression (\eg, $BA + B\bar{A}C$ in the bottom row of~\Cref{supp_fig:ilp}) that represents precondition.
Finally, we build the subtask graph by consolidating the preconditions of all subtasks.

\subsection{Subtask Graph Inference from Real-world Data}
\label{supp_subsec:ilp_ours}
\paragraph{Layer-wise Precondition Inference.}
One major problem of inferring the precondition independently for each subtask is the possibility of forming a \emph{cycle} in the resulting subtask graph, which leads to a causality paradox (\ie, subtask A is a precondition of subtask B and subtask B is a precondition of subtask A).
To avoid this problem, we perform precondition inference in a \emph{layer-wise} fashion similar to~\citet{sohn-iclr20}.

To this end, we first infer the layer of each subtask from the (noise reduced) subtask state labels $\{\stateset^i\}_{i=1}^{|\mb{V}_\tau|}$ for the task $\tau$.
\citet{sohn-iclr20}~infer the layer of each subtask by finding the minimal set of subtask completions to perfectly discriminate the eligibility of a subtask.
However, this approach is not applicable to our setting due to a lack of data points with ineligible subtasks; \ie, we can perfectly discriminate the eligibility by predicting a subtask to be \emph{always} eligible.
Instead, we propose to extract the parent-child relationship from the subtask state labels $\stateset$.
Intuitively speaking, we consider subtask $n$ to be the ancestor of subtask $m$ if subtask $n$ (almost) always precedes subtask $m$ in the videos, and assign at least one greater depth to subtask $m$ than subtask $n$.
Specifically, we count the (long-term) transitions between all pairs of subtasks and compute the \emph{transition purity} as follows:
\begin{align}
    \text{purity}_{n\rightarrow m} = \frac{\text{\# occurences subtask $n$ preceds subtask $m$ in the video}}{\text{\# occurences subtask $n$ and subtask $m$ appears together in the video}}
\end{align}
We consider subtask $n$ to be the ancestor of subtask $m$ if the transition purity is larger than a threshold $\delta$  (\ie, $\text{purity}_{n\rightarrow m} > \delta)$.
Intuitively, when the data has higher noise, we should use lower $\delta$. We used $\delta=0.96$ for ProceL and $\delta=0.55$ for CrossTask in the experiment.
Note that this is only a \emph{necessary} condition, and it cannot guarantee to extract \emph{all} of the parent-child relationships, especially when the precondition involves \andorgraph{OR} (\( | \)) relationship.
Thus, we use this information only for deciding the layer of each subtask and do not use it for inferring the precondition.

After each subtask is assigned to its depth, we perform precondition inference at each depth in order of increasing depth.
By definition, the subtasks at depth$=0$ do not have any precondition.
When inferring the precondition of a subtask in layer $l$, we use the completion of subtasks in depth $1, \ldots, (l-1)$.
This ensures that the edge in the subtask graph is formed from the lower depth to the higher depth, which prevents the cycle.

\paragraph{Recency Weighting.}
To further improve the graph generation, we propose to take the temporal information into account.
In fact, the conventional ILP does not need to take the time step into account since it assumes eligibility data to be available at every time step. 
However, in our case, we are only given a single data point with positive eligibility per video clip per subtask, where incorporating the temporal information can be very helpful.
Motivated by this, we propose to assign the weight to each data sample according to the \textit{recency}; \ie, we assign higher weight if a subtask has become eligible more recently.
We first rewrite~\Cref{eq:ilp-new}~as follows:
\begin{align}
\hat{\pfunc}
&=\argmax_{\pfunc} \left\{P\left(e_n=1 | \pfunc(\mb{c})=1\right) - \alpha C\left(\pfunc\right)\right\}\\
&\simeq \argmax_{\pfunc}\frac{
    \frac{1}{|\ilpdata_n|} \sum_{j=1}^{|\ilpdata_n|}
    {
        \mbb{I}\left(
            \pfunc(\mb{c}^{j})=1, e^{j}_{n}=1
        \right)
    }
}{
    \frac{1}{2^{N}} \sum_{\mb{c}}
    {
        \mbb{I}\left(
            \pfunc(\mb{c})=1
        \right)
    }
} - \alpha C\left(\pfunc\right)\\
&=\argmax_{\pfunc}\frac{
    \frac{1}{|\mb{X}_\tau|} \sum_{i=1}^{|\mb{X}_\tau|}
    {
        \frac{1}{T}
        \sum_{t=1}^{T}
        {
            \mbb{I}
            \left(
                \pfunc(\mb{c}^{i}_{t})=1, e^{i}_{t, n}=1
            \right)
        }
    }
}{
    \frac{1}{2^{N}} \sum_{\mb{c}}
    {
        \mbb{I}\left(
            \pfunc(\mb{c})=1
        \right)
    }
} - \alpha C\left(\pfunc\right)\label{suppeq:ilp-rewrite}
\end{align}
Then, we add the recency weight ${\color{blue}w_{t, n}}$ and modify~\Cref{suppeq:ilp-rewrite} as follows: 
\begin{align}
\hat{\pfunc}
&\simeq \argmax_{\pfunc}\frac{
    \frac{1}{|\mb{X}_\tau|} \sum_{i=1}^{|\mb{X}_\tau|}
    {
        \frac{1}{T}
        \sum_{t=1}^{T}
        {
            {\color{blue}w_{t, n}} \ \mbb{I}
            \left(
                \pfunc(\mb{c}^{i}_{t})=1, e^{i}_{t,n}=1
            \right)
        }
    }
}{
    \frac{1}{2^{N}} \sum_{\mb{c}}
    {
        \mbb{I}\left(
            \pfunc(\mb{c})=1
        \right)
    }
} - \alpha C\left(\pfunc\right)\label{suppeq:ilp-recency},
\end{align}
where 
\begin{align}
    w_{t, n} = \max(0.1, \lambda ^ {t_n-t}),
\end{align}
$0<\lambda<1$ is the discount factor, $t_n$ is the time step when the precondition for subtask $n$ became satisfied.

\paragraph{Hyperparameters.}
We used $\alpha=0.2$ and $\lambda=0.7$ in our experiments.

\subsection{Task-level Graph Generation Results}
\label{supp_subsec:task_level_graph_generation_result}

We present graph generation metrics for each task separately in \Cref{supptab:graph_evaluation}.

\begin{table*}[ht]
    \setlength{\tabcolsep}{4.2pt}
    \centering
    \footnotesize
    \caption{\textbf{Task-level Graph Generation Result in ProceL~\cite{elhamifar-iccv19}.} Task label indexes (a-l) are identical to~\Cref{supptab:frame_evaluation}.}
    \begin{tabular}{cl|cccccccccccc|c}
        \toprule  
        Metric & Method & (a) & (b) & (c) & (d) & (e) & (f) & (g) & (h) & (i) & (j) & (k) & (l) & Avg\\
        \midrule
        \multirow{5}{*}{\rotatebox[origin=c]{90}{Accuracy}}
                                                & proScript & 54.69 & 55.56 & 62.50 & 63.54 & 53.57 & 62.50 & 58.93 & 57.14 & 57.14 & 54.17 & 55.00 & 55.21 & 57.50 \\
                                                & MSGI & 50.00 & 55.56 & 53.12 & 55.21 & 48.81 & 52.50 & 58.93 & 55.36 & 51.79 & 58.33 & 50.00 & 61.11 & 54.23 \\
                                                & \msgip & 57.03 & 66.67 & 53.12 & 60.42 & 57.74 & 68.75 & 61.61 & 67.86 & 64.29 & 63.54 & 58.44 & 64.58 & 62.00 \\
                                                & \ourGraph & 73.44 & 80.56 & 81.25 & 84.38 & 88.69 & 90.00 & 76.79 & 98.21 & 85.71 & 81.25 & 72.50 & 66.67 & 81.62 \\
                                                & \textbf{\ourMethodName~(Ours)} & 87.50 & 79.17 & 90.62 & 84.38 & 83.33 & 85.00 & 67.86 & 100.00 & 87.50 & 83.33 & 82.50 & 66.67 & \textbf{83.16} \\
        \midrule
        \multirow{5}{*}{\rotatebox[origin=c]{90}{\ourGraphMetric}}
                                                & proScript & 72.08 & 65.03 & 57.14 & 60.61 & 64.52 & 70.00 & 65.38 & 62.09 & 58.24 & 47.73 & 60.00 & 65.28 & 62.34 \\
                                                & MSGI & 83.33 & 66.34 & 69.64 & 58.33 & 60.00 & 61.11 & 57.69 & 50.55 & 59.89 & 67.42 & 74.44 & 66.67 & 64.62 \\
                                                & \msgip & 77.92 & 70.59 & 69.64 & 62.12 & 64.76 & 82.22 & 69.23 & 84.07 & 72.53 & 65.91 & 73.33 & 76.39 & 72.39 \\
                                                & \ourGraph & 87.92 & 92.48 & 91.07 & 93.18 & 86.90 & 90.00 & 82.97 & 93.41 & 89.56 & 90.15 & 77.78 & 84.72 & 88.35 \\
                                                & \textbf{\ourMethodName~(Ours)} & 95.83 & 88.89 & 92.86 & 93.18 & 90.00 & 92.22 & 89.01 & 100.00 & 90.11 & 87.12 & 83.33 & 76.39 & \textbf{89.91} \\
        \midrule
        \multirow{5}{*}{\rotatebox[origin=c]{90}{\ourDataMetric}}
                                                & proScript & 26.20 & 9.30 & 51.27 & 30.12 & 28.70 & 52.29 & 25.74 & 47.75 & 34.12 & 15.55 & 48.03 & 76.40 & 37.12 \\
                                                & MSGI & \multicolumn{12}{c|}{N/A} & N/A \\
                                                & \msgip & 71.62 & 81.62 & N/A & 67.80 & 62.18 & 73.53 & 80.94 & 88.41 & 87.51 & 73.16 & 73.54 & 76.60 & 76.08 \\
                                                & \ourGraph & 98.14 & 99.14 & 99.39 & 98.78 & 100.00 & 98.88 & 94.41 & 96.71 & 100.00 & 97.40 & 95.58 & 95.93 & 97.86 \\
                                                & \textbf{\ourMethodName~(Ours)} & 100.00 & 98.60 & 98.52 & 98.78 & 98.57 & 98.55 & 96.65 & 99.43 & 98.26 & 97.19 & 97.48 & 97.53 & \textbf{98.30} \\
        \bottomrule
    \end{tabular}
    \label{supptab:graph_evaluation}
\end{table*}

\subsection{Generated Graphs}
\label{supp_subsec:graph_generation}
\begin{figure}[t]
    \centering
    \begin{subfigure}{\linewidth}
        \centering
        \begin{subfigure}{\linewidth}
            \centering
            \includegraphics[width=0.6\linewidth]{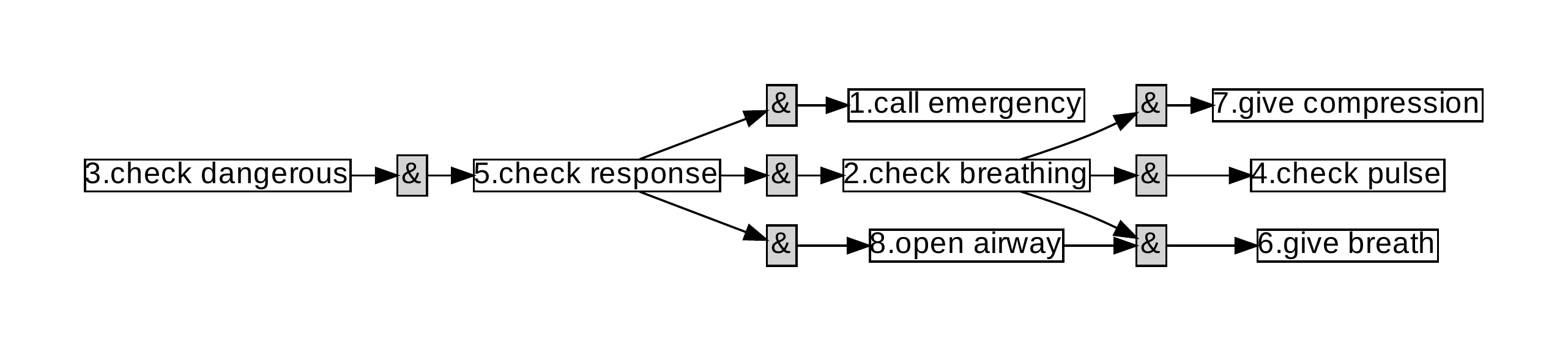}
            \vspace{-10px}
            \caption*{1. Human-annotated graph}
        \end{subfigure}
        \begin{subfigure}{\linewidth}
            \centering
            \includegraphics[width=0.6\linewidth]{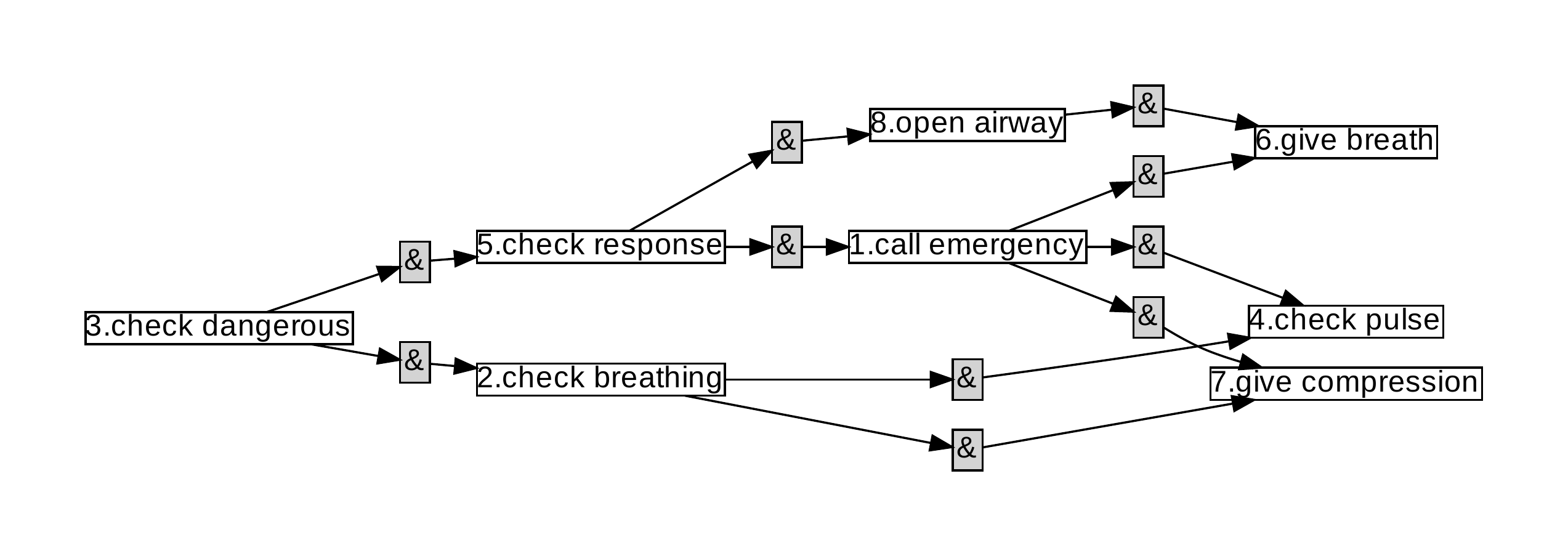}
            \vspace{-10px}
            \caption*{2. \ourGraphFull}
        \end{subfigure}
        \begin{subfigure}{\linewidth}
            \centering
            \includegraphics[width=0.7\linewidth]{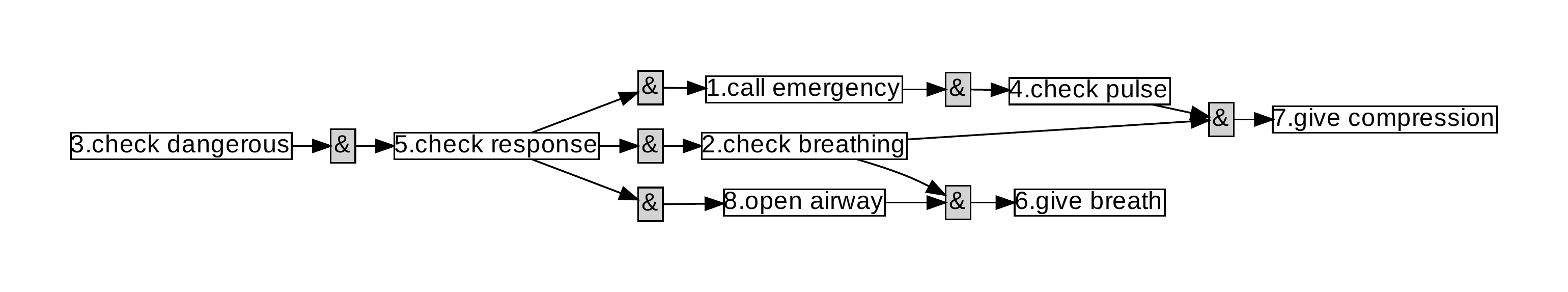}
            \vspace{-10px}
            \caption*{3. \ourMethodName}
        \end{subfigure}
        \caption{Perform CPR}
        \label{supp_subfig:perform_cpr}
    \end{subfigure}
    \begin{subfigure}{\linewidth}
        \centering
        \begin{subfigure}{\linewidth}
            \centering
            \includegraphics[width=0.75\linewidth]{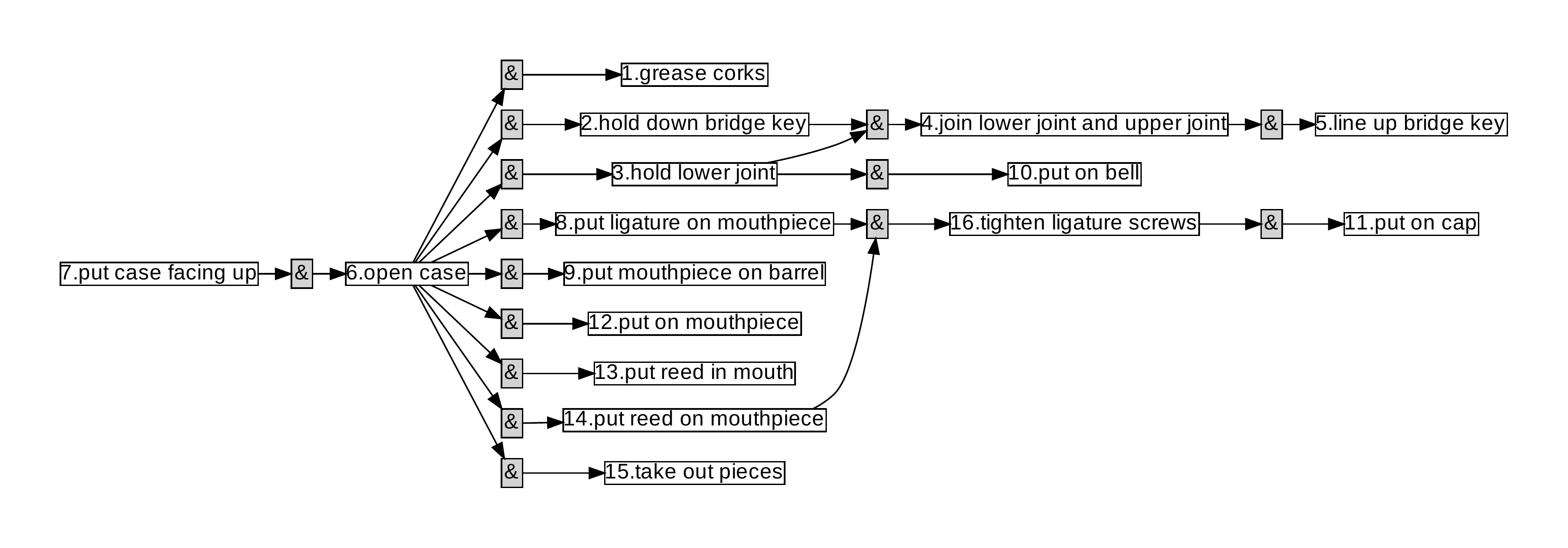}
            \vspace{-10px}
            \caption*{1. Human-annotated graph}
        \end{subfigure}
        \begin{subfigure}{\linewidth}
            \centering
            \includegraphics[width=0.75\linewidth]{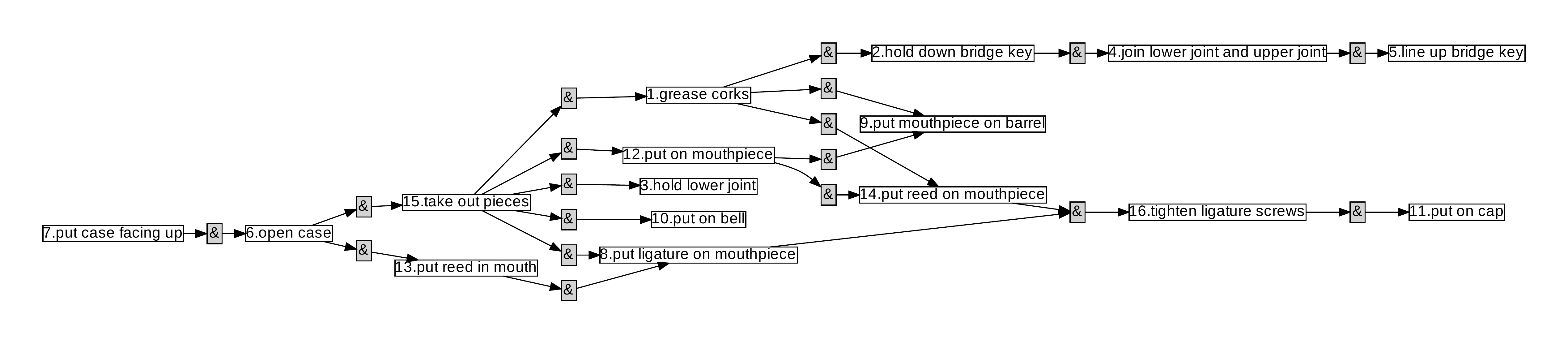}
            \vspace{-10px}
            \caption*{2. \ourGraphFull}
        \end{subfigure}
        \begin{subfigure}{\linewidth}
            \centering
            \includegraphics[width=0.75\linewidth]{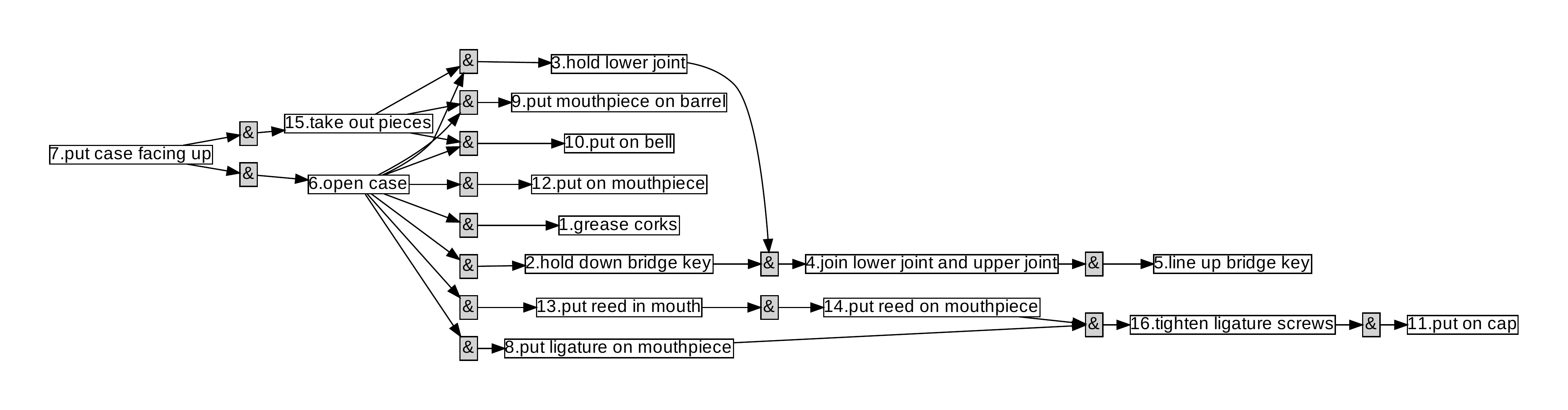}
            \vspace{-10px}
            \caption*{3. \ourMethodName}
        \end{subfigure}
        \caption{Assemble Clarinet}
        \label{supp_subfig:assemble_clarinet}
    \end{subfigure}
    \caption{\textbf{Ground Truth and Generated Graphs.} We plot the subtask graphs for (a) Perform CPR and (b) Assemble Clarinet. %
    In each subfigure, the three graphs correspond to 1. Human-annotated graph, 2. \ourGraphFull, and 3. \ourMethodName~from the ProceL dataset, respectively.
    }
    \label{supp_fig:generated_graph_examples}
\end{figure}
To evaluate the performance of our graph generation approach, we hand-annotate subtask graphs for each task in the ProceL dataset, and we plot the subtask graphs for two tasks in~\Cref{supp_fig:generated_graph_examples}.
In addition to the hand-annotated graphs, we also plot the subtask graphs generated from~\ourMethodName~and~\ourGraphFull. %
When we compare the ground-truth with~\ourMethodName, our predictions closely resemble the ground-truth graphs, compared with~\ourGraphFull. 
These results show that our subtask state prediction improves subtask labels in the data, which leads to more accurate generated graphs.

\section{Next Step Prediction with Subtask Graph}
\label{supp_sec:next_step_pred}

\subsection{Details about Next Step Prediction with Graphs}
\label{supp_subsec:next_step_pred_details}

From the subtask label $\statevec_{t}$ and predicted subtask state $\hat{\statevec}_{t}$, we first obtain the subtask completion $\mb{c}_t$ by checking whether each subtask state is $\statec$.
Then, we compute the subtask eligibility $\mb{e}_t$ from the completion $\mb{c}_t$ and subtask graph $G$ as $\mb{e}_t = \pfuncarg{G}(\mb{c}_t)$.
When predicting the next step subtask, we exploit the fact that a subtask can be completed only if 1) it is eligible and 2) it is incomplete.
Thus, we compute the subtask prediction mask $\mb{m}_t$ as $\mb{m}_t = \mb{e}_t \odot (1 - \mb{c}_t)$, where $\odot$ denotes element-wise multiplication.
Lastly, among the eligible and incomplete subtasks, we assign a higher probability if a subtask has been eligible more recently: $p_{t+1}[n] \propto m_t[n] \cdot \rho ^ {\Delta t^\text{elig}_n}$,
where $p_t[n]$ is the probability that $n$-th subtask is (or will be) completed at time $t$, $\Delta t^\text{elig}_n$ is the time steps elapsed since the $n$-th subtask has been eligible, and $0 < \rho < 1$ is the discount factor. We used $\rho=0.9$ in the experiment.

\subsection{Training Details}
\label{supp_subsec:training_details_next_step_pred_result}

We apply our subtask graph generation method to the next subtask prediction task in~\Cref{subsec:experiment_next_step_prediction}.
For both ProceL~\cite{elhamifar-iccv19} and CrossTask~\cite{zhukov-cvpr19}, we first convert all videos to 30 fps, obtain the verb phrases, and extract CLIP features, following~\Cref{supp_subsec:training_details_subtask_state_prediction}.
We split 15\% of the data as the test set. %
During training, we randomly select a subtask and feed the data up to the selected subtask with subsampling at least three frames from each subtask region but no more than 190 frames in total for all models.
For evaluation, we provide all previous subtasks in the dataset and sample 3 frames per subtask in an equidistant manner (without any random sampling).
All the other settings are the same as subtask state prediction.

\subsection{Task-level Next Step Prediction Results}
\label{supp_subsec:task_level_next_step_pred_result}

We share task-level next step prediction results in~\Cref{supptab:next_subtask_pred}.
All the method labels correspond to~\Cref{tab:next_task_pred}.

\begin{table*}[t]
    \setlength{\tabcolsep}{3.0pt}
    \centering
    \footnotesize
    \caption{\textbf{Task-level Next Step Prediction Results.} We perform next subtask prediction on ProceL~\cite{elhamifar-iccv19} and CrossTask~\cite{zhukov-cvpr19} datasets and measure the accuracy(\%). Task label indexes (a-l) in ProceL task are the same as~\Cref{supptab:frame_evaluation}. Task labels (i-x) in CrossTask task are (i) Add Oil to Your Car, (ii) Build Simple Floating Shelves, (iii) Change a Tire, (iv) Grill Steak, (v) Jack Up a Car, (vi) Make a Latte, (vii) Make Banana Ice Cream, (viii) Make Bread and Butter Pickles, (ix) Make French Strawberry Cake, (x) Make French Toast, (xi) Make Irish Coffee, (xii) Make Jello Shots, (xiii) Make Kerala Fish Curry, (xiv) Make Kimchi Fried Rice, (xv) Make Lemonade, (xvi) Make Meringue, (xvii) Make Pancakes and (xviii) Make Taco Salad.}
    \begin{tabular}{r|l|cccccccccccc|c}
    \toprule  
                    & \textbf{Model} & (a) & (b) & (c) & (d) & (e) & (f) & (g) & (h) & (i) & (j) & (k) & (l) & Avg \\ \midrule
    \multirow{7}{*}{\rotatebox[origin=c]{90}{\textbf{ProceL}}} & STAM~\cite{sharir-arxiv21} & 23.61 & 23.81 & 21.43 & 46.88 & 25.00 & 63.04 & 57.14 & 34.29 & 4.88 & 24.53 & 27.50 & 6.25 & 29.86 \\ 
                                     & ViViT~\cite{arnab-iccv21} & 26.39 & 20.24 & 19.05 & 50.00 & 25.00 & 28.26 & 55.36 & 40.00 & 4.88 & 32.08 & 22.50 & 0.00 & 26.98 \\ \cline{2-15}
                                     & proScript~\cite{sakaguchi-emnlpf21} & 15.57 & 10.34 & 37.50 & 36.17 & 10.62 & 32.76 & 21.05 & 28.06 & 13.19 & 8.33 & 12.70 & 0.00 & 18.86 \\
                                     & MSGI~\cite{sohn-iclr20} & 10.66 & 13.10 & 32.81 & 25.53 & 7.96 & 25.86 & 14.74 & 24.46 & 13.19 & 13.54 & 7.94 & 19.23 & 17.42 \\
                                     & \msgip & 22.13 & 36.55 & 32.81 & 26.60 & 25.66 & 37.93 & 20.00 & 34.53 & 16.48 & 27.08 & 7.94 & 30.77 & 26.54 \\
                                     & \ourGraph & 31.97 & 56.55 & 43.75 & 73.40 & 51.33 & 51.72 & 46.32 & 69.06 & 50.55 & 48.96 & 30.16 & 26.92 & 48.39 \\
                                     & \textbf{\ourMethodName~(Ours)} & 40.16 & 51.72 & 56.25 & 73.40 & 62.83 & 68.97 & 44.21 & 69.06 & 59.34 & 48.96 & 39.68 & 50.00 & \textbf{55.38} \\
    \bottomrule
    \end{tabular}
    \setlength{\tabcolsep}{6.0pt}
    \begin{tabular}{r|l|ccccccccc|c}
                                   & \multirow{2}{*}{\textbf{Model}} & (i) & (ii) & (iii) & (iv) & (v) & (vi) & (vii) & (viii) & (ix) & \multirow{2}{*}{Avg}\\
                                   &                                 & (x) & (xi) & (xii) & (xiii) & (xiv) & (xv) & (xvi) & (xvii) & (xviii) & \\     \midrule
    \multirow{14}{*}{\rotatebox[origin=c]{90}{\textbf{CrossTask}}} 
                                        & \multirow{2}{*}{STAM~\cite{sharir-arxiv21}} & 30.77 & 58.82 & 48.21 & 38.82 & 27.27 & 26.98 & 34.00 & 21.31 & 15.69 & \multirow{2}{*}{40.17}\\ 
                                        &                                            & 47.91 & 78.57 & 48.94 & 34.83 & 38.60 & 36.90 & 54.93 & 42.86 & 37.63 & \\
                                        & \multirow{2}{*}{ViViT~\cite{arnab-iccv21}} & 34.62 & 43.14 & 49.11 & 34.12 & 63.64 & 34.92 & 60.00 & 13.11 & 11.76 & \multirow{2}{*}{41.96}\\ 
                                        &                                            & 48.37 & 82.14 & 50.00 & 32.58 & 38.60 & 35.71 & 46.48 & 43.57 & 33.33 & \\\cline{2-12}
                                        & \multirow{2}{*}{proScript~\cite{sakaguchi-emnlpf21}} & 21.37 & 50.65 & 37.93 & 20.07 & 90.62 & 33.33 & 48.15 & 22.52 & 34.12 & \multirow{2}{*}{36.78}\\
                                        &                                         & 33.44 & 45.54 & 33.78 & 26.36 & 33.72 & 38.89 & 42.57 & 24.49 & 24.40 & \\
                                        & \multirow{2}{*}{MSGI~\cite{sohn-iclr20}}   & 30.77 & 32.47 & 23.45 & 14.60 & 71.88 & 33.33 & 39.51 & 18.02 & 18.82 & \multirow{2}{*}{32.31}\\
                                        &                                            & 22.51 & 33.04 & 36.49 & 33.33 & 45.35 & 32.54 & 34.65 & 34.69 & 26.19 & \\
                                        & \multirow{2}{*}{\msgip}              & 30.77 & 32.47 & 22.76 & 14.60 & 71.88 & 33.33 & 39.51 & 27.93 & 18.82 & \multirow{2}{*}{32.72}\\
                                        &                                            & 20.58 & 33.04 & 36.49 & 33.33 & 45.35 & 32.54 & 34.65 & 34.69 & 26.19 & \\
                                        & \multirow{2}{*}{\ourGraph}                 & 67.52 & 72.73 & 64.83 & 45.99 & 90.62 & 44.74 & 48.15 & 32.43 & 37.65 & \multirow{2}{*}{53.39}\\
                                        &                                            & 63.99 & 50.89 & 52.03 & 47.29 & 48.84 & 36.51 & 58.42 & 63.78 & 34.52 & \\
                                        & \multirow{2}{*}{\textbf{\ourMethodName~(Ours)}} & 67.52 & 72.73 & 68.28 & 41.24 & 90.62 & 46.49 & 48.15 & 32.43 & 37.65 & \multirow{2}{*}{\textbf{54.42}}  \\ 
                                        &                                             & 63.99 & 55.36 & 57.43 & 48.06 & 48.84 & 46.03 & 58.42 & 64.80 & 31.55 & \\
    \bottomrule
    \end{tabular}
    \label{supptab:next_subtask_pred}
\end{table*}

\end{document}